\documentclass[10pt,twocolumn,letterpaper]{article}

\usepackage{wacv}
\usepackage{times}

\usepackage{epsfig}
\usepackage{amsmath}
\usepackage{amssymb}

\usepackage[accsupp]{axessibility} 

\usepackage{subcaption}
\usepackage{commath}
\usepackage{xcolor}
\usepackage{multirow, booktabs, hhline}
\usepackage[normalem]{ulem}

\usepackage{enumitem}

\usepackage{array}
\newcommand{\PreserveBackslash}[1]{\let\temp=\\#1\let\\=\temp}
\newcolumntype{C}[1]{>{\PreserveBackslash\centering}p{#1}}
\newcolumntype{R}[1]{>{\PreserveBackslash\raggedleft}p{#1}}
\newcolumntype{L}[1]{>{\PreserveBackslash\raggedright}p{#1}}

\def\wacvPaperID{388} 

\wacvfinalcopy 

\ifwacvfinal
\usepackage[breaklinks=true,colorlinks,bookmarks=false]{hyperref} 
\else
\usepackage[pagebackref=true,breaklinks=true,colorlinks,bookmarks=false]{hyperref}
\fi

\begin{document}


\title{Quantified Facial Expressiveness for Affective Behavior Analytics}

\author{Md Taufeeq Uddin\\
University of South Florida, Tampa, FL, US\\
{\tt\small mdtaufeeq@usf.edu}
\and
Shaun Canavan\\
University of South Florida, Tampa, FL, US\\
{\tt\small scanavan@usf.edu}
}

\maketitle

\begin{abstract}

The quantified measurement of facial expressiveness is crucial to analyze human affective behavior at scale. Unfortunately, methods for expressiveness quantification at the video frame-level are largely unexplored, unlike the study of discrete expression. In this work, we propose an algorithm that quantifies facial expressiveness using a bounded, continuous expressiveness score using multimodal facial features, such as action units (AUs), landmarks, head pose, and gaze. The proposed algorithm more heavily weights AUs with high intensities and large temporal changes. The proposed algorithm can compute the expressiveness in terms of discrete expression, and can be used to perform tasks including facial behavior tracking and subjectivity quantification in context. Our results on benchmark datasets show the proposed algorithm is effective in terms of capturing temporal changes and expressiveness, measuring subjective differences in context, and extracting useful insight.

\end{abstract}

\section{Introduction}
\label{sec:intro}



\begin{figure*}
\centering
  \centering
  \includegraphics[width=.75\linewidth]{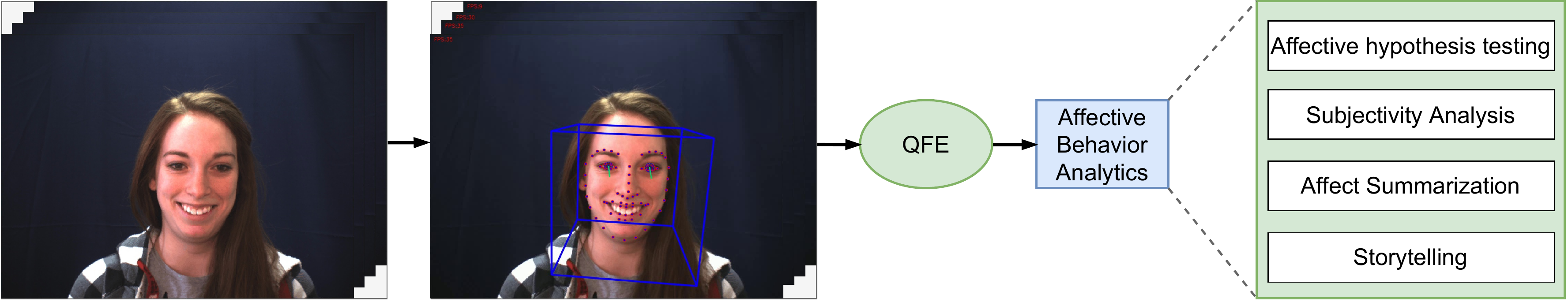} 
  \vspace{-2mm}
  \caption{Workflow of the proposed method. Given affect videos, multimodal facial features are extracted to use as input to the proposed QFE algorithm. Then, the computed expressiveness score is used along with other modalities such as context to perform affective behavior analytics to perform varied tasks. These tasks include, but are not limited to, enhancing emotional signal analysis, affective hypotheses testing, subjective difference analysis, summarizing affect data, and tell engaging stories.}
\label{fig:sampSeq}
\vspace{-5mm}
\end{figure*}
Affective data analytics can be a powerful tool to explore expressions within context to discover underlying patterns and relationships between expressions and other variables of interest (e.g., EEG data \cite{koelstra2011deap}). It can be especially useful since there are two opposing theories about emotional expressions \cite{siegel2018emotion}, namely the classical view of emotion, and the theory of constructed emotion. The classical view of emotion states that emotions are universal among humans, whereas the theory of constructed emotion states that emotions come from the complex dynamics of humans and context \cite{barrett2017functionalism}. It has also been shown that expressiveness is subjective and happens at different frequencies and intensities \cite{waterman1990personal}. Tools for analyzing expressions allow for insight into affective data and how it relates to each opposing theory. 

While expressiveness has been extensively studied in psychology \cite{balswick1977differences,dunsmore1997does,friedman1980understanding,roberts1996empathy}, fewer works appear in affective computing. With the increase in large-scale emotion-based datasets \cite{cheng20184dfab,zhang2016multimodal}, the current manual approach to annotating expressiveness \cite{lin2019context} is not scalable. An automated approach is needed to objectively, and quickly analyze facial images to facilitate further advances in affective computing, especially as the need for data grows with deep learning approaches to expression \cite{Wang_2020_CVPR} and emotion \cite{mittal2020emoticon} recognition.


The difficulties with manual annotation and the importance of emotional expressiveness \cite{cassidy1992family} motivates us to quantify facial expressiveness within context (i.e. external stimuli). This can be useful for more objective scientific studies with affective data, as well as quantitatively evaluating the differences in expressiveness between people. As more context-aware affect models \cite{lee2019context} are developed, a better understanding of context can also be useful. Considering this, we propose to analyze expressiveness as it is related to the context (i.e., external stimuli are used to elicit expressions, which occur at different intensities). We investigate two publicly available datasets, namely DISFA \cite{mavadati2013disfa} and BioVid Pain \cite{walter2013biovid} datasets. We find that context influences expressiveness and there is a subjective difference in the intensity and frequency of said expressiveness. The main contributions of this work are detailed below.
\vspace{-3mm}
\begin{enumerate}
    \setlength{\itemsep}{1pt}
    \setlength{\parskip}{0pt}
    \setlength{\parsep}{0pt}
    \item A quantified approach to the analysis of facial expressiveness is proposed (Fig. \ref{fig:sampSeq}). It is bounded by a lower and upper limit of expressiveness, which allows us to more objectively compare different data. 
    
    \item Detailed analysis of the relationship between context and expressiveness is given on two publicly available datasets. Our results suggest that different context can impact the overall expressiveness of subjects. A Granger causality-based hypothesis between facial expressiveness and temporal context is also tested.
    
    \item The subjective differences in expressiveness are demonstrated using the proposed, bounded, quantitative approach. We show that given the same context (i.e. the subjects are introduced to the same external stimuli), different subjects have different intensity and frequency of expressiveness.
    
    \item We demonstrate how the proposed algorithm can be used to analyze, summarize, and interpret human affective behavior exploiting affect videos and relevant information to augment affective computer vision. 
    
\end{enumerate}
\vspace{-3mm}
\section{Related Works}
\label{sec:relWorks}
\vspace{-3mm}
Many works in psychology have studied different types of expressiveness including personal \cite{waterman1990personal}, family \cite{halberstadt1999family}, and nonverbal \cite{friedman1980understanding}. Ogren and Johnson \cite{ogren2021primary} found that the expressiveness of the primary caregiver of children strongly relates to their understanding of emotion. Ludwisowski et al. \cite{ludwikowski2018explaining} investigated the relationship of gender and expressiveness, with a specific focus on how it can explain different gender interests. They found females were more expressive than males, which had a chain effect that impacted artistic interests.  Self-report is generally considered accurate, however, subjects may not be truthful on them \cite{fuentes2017systematic}. Although psychologists rely on self-report \cite{barrett2004feelings}, having an automatic approach to analyze the emotional expressiveness of a subject would offer a fast, objective alternative.

Over the last few decades, researchers studied affect in numerous ways such as categorical (happy, sad) and dimensional (valence, arousal) \cite{grandjean2008conscious}. Normally, data were annotated by subjects (self-report) or an observer. One of the limitations of the dimensional approach is that it provides comparatively generic information such as unpleasant to pleasant, which is often used in sentiment modeling \cite{barriere2020improving}. On the other hand, categorical models use classes such as happy, surprise, or sad \cite{liu2021point}. Also, many studies focus on the presence or absence of each class, not including the intensity of the expression \cite{li2020deep}. That being said, there are recent works focused on these limitations. For example, Lin et al. \cite{lin2019context, lin2020toward}, and Lei et al. \cite{lei2020emotion} measured the facial expressiveness at the video sequence level using human annotators. Although these results are encouraging, it does have the limitation that subjective human ratings need to be collected, which is time-consuming, and lack of expressiveness details. Note that the expressiveness is not uniform throughout the sequence. As pointed out by Gunes et al. \cite{gunes2011emotion}, a single label (annotation) may not capture the complexity of expressions. Hence, we need methods that can measure expressiveness at granular level in multiple dimensions (expressions are likely to be mixed \cite{blank2020emotional} such as joy, happy, celebration). 

Uddin and Canavan \cite{uddin2021quantified} proposed TED to quantify facial expressiveness. While encouraging, there are some limitations to this approach that motivate our current work into quantified facial expressiveness. Their proposed approach can't measure dynamic changes (e.g. gaze) properly, and the quantification is unbounded and biased towards the number of action units, which makes the comparison of different expression data infeasible. Our proposed algorithm extends state of the art by removing the need for human annotations and providing a fast and objective measure of affective expressiveness that can be used to compare multiple datasets. The proposed approach can be used on various types of expressiveness including but not limited to mixed, complex, and simultaneous.

\vspace{-3mm}
\section{Quantified Facial Expressiveness}
\vspace{-2mm}



\subsection{Quantified Facial Expressiveness Algorithm}


There are two major components of facial expressiveness: spatial (static) and temporal (dynamic). Spatial expressiveness is observed in a static video frame in a given moment in time. This expressiveness can be captured from the intensities of facial AUs given that AUs have well-defined meaning based on the classical view of emotion \cite{ekman1997face, kring2007facial}. AUs are also associated with individual expressiveness, personality, stimuli, and self-report \cite{ekman1976measuring}. Here, we compute a spatial, continuous expressiveness score for a given frame bounded by a lower and upper limit by 
\begin{equation}
\label{eqn:staticExprv}
\sigma = \frac{\lambda}{n [\exp (1) - 1]} \sum \limits _{i=1} ^{n} \Big [\exp \Big(\frac{x_i}{x_{max}}\Big) - 1 \Big]
\end{equation}
where $x_i$ is a vector containing the intensities of AUs of interest, $n$ is the length of the vector, and $x_{max}$ is the maximum possible intensity of the AUs. By convention, AUs are coded in between $[0, 5]$ where $0$ indicates absence of the AU, and $5$ indicate maximum activated AU intensity. The motivation behind Eqn. \ref{eqn:staticExprv} is to more heavily weight the active AUs, while bounding the spatial expressiveness score in between $[0, \lambda]$, where $\lambda$ is a constant multiplier. It is important to note that Eqn. \ref{eqn:staticExprv} can only capture the spatial (static) expressiveness. Hence, to capture temporal expressiveness, other essential modalities such as facial landmarks, head pose, and eye gaze are exploited. These modalities don't have intensity as AUs do and are generally represented by coordinates in $2D, 3D$ space, or orientation and rotation. Considering this, we track temporal expressiveness from these modalities using the following set of equations. First, we measure the relative change by computing the velocity for consecutive frames (Eqn. \ref{eqn:vel}).

\begin{equation}
    \Delta v = \frac{\Delta x}{\Delta t}
    \label{eqn:vel}
\end{equation}

Where $\Delta x$ and $\Delta t$ represent the change in corresponding values between two frames, and interval between the frames, respectively. As we want to more heavily weight the location where major change happens, and approximate the information in the neighboring frames, we approximate the temporal expressiveness for each modality using the Taylor series approximation of the following exponential function: $e^y -1$, in our case, $y = \Delta v$ \footnote{$e^y - 1 = e^{\Delta v} - 1 = \sum_{m=1}^{\infty} \frac{\Delta v^m}{m!} = \Delta v + \frac{\Delta v^{2}}{2!} + \frac{\Delta v^{3}}{3!} + \dots$}. Note that $e^{\Delta v} - 1$ is bounded by $[0, 1.718]$  when $0 \leq \Delta v \leq 1$, which is useful to get a lower and upper bounded temporal expressiveness score. Hence, for a given modality, for each pair of points, we approximate the temporal expressiveness using Eqn. \ref{eqn:tempoExprv}, where $n$ and $m$ are the length of facial feature vector, and the order to which the approximation is performed, respectively. 
\begin{equation}
\label{eqn:tempoExprv}
    t_{exp} = \sum \limits _{j=1} ^{n} [\exp(\Delta v) - 1] = \sum \limits _{j=1} ^{n}  \sum \limits _{m=1} ^{\infty} \frac{\Delta v^m}{m!}
\end{equation}
It is then scaled to $[0, 1]$ (Eqn. \ref{eqn:tempoExprvNorm}), in which $\Delta_{max} = 1$ given the feature vectors are scaled between $[0, 1]$. 
\begin{equation}
\label{eqn:tempoExprvNorm}
     \delta = \frac{t_{\exp}}{n * [\exp(\Delta_{max}) - 1]} 
\end{equation}
From Eqns. \ref{eqn:staticExprv} and \ref{eqn:tempoExprvNorm}, spatial expressiveness, $\sigma$, and temporal expressiveness, $\delta$, are in between $[0, \lambda]$, and $[0, 1]$.

\vspace{-4mm}
\subsubsection{\textbf{Combining spatial and temporal expressiveness}} 
\vspace{-3mm}
\textbf{Approach $1$.} We hypothesize that $\sigma$ is the main source of expressiveness following literature of the classical view of emotion \cite{ekman1976measuring}, and $\delta$ is the auxiliary source of expressiveness. Hence, to obtain the quantified facial expressiveness (\textbf{QFE}) score ($\tau$) treating $\sigma$ as the essential source of expressiveness, we combine $\sigma$ and $\delta$ by
\begin{equation}
\label{eqn:QFE_scr}
\tau = \sigma * \Big [1 + \frac{1}{n_{mod}} \sum \limits _{k=1} ^{n_{mod}} \lambda_{k} \delta_{k} \Big].
\end{equation}
Here, $\lambda_k$ represents the weight parameter for a given temporal modality and $n_{mod}$ represents the number of modalities, which are needed to compute the weighted mean of the temporal modalities. Hence, $\tau$ represents the QFE score for a given face for a given moment in time. Notice that depending on the $\lambda_k$, we can have $\tau$ bounded in between $[0, n_b \lambda]$, where $n_b$ is a scalar. For instance, from Eqn. \ref{eqn:QFE_scr}, if we set $\lambda_{k} = 1$, then the QFE score is $0 \leq \tau \leq 2 \lambda$.

\textbf{Approach $2$}. $\sigma$ and $\delta$ can be combined using the weighted combination with an additional adjustment term as offset, i.e. $\tau_{wc} = w_i\sigma + w_{i+1} \delta + \epsilon$, where $w_i$ and $w_{i+1}$ are the weights and $\epsilon$ is an adjustment term. There could be scenarios where this formulation could be relevant: i) both spatial and temporal modalities are equally important; ii) temporal expressiveness is more crucial than spatial expressiveness. For instance, in the case of student engagement, autism spectrum disorder, or driver behavior studies, eye gaze could be more relevant than other modalities including AUs \cite{fabiano2020gaze,srivastava2020recognizing}. In these scenarios, $\tau_{wc}$ maybe more effective and can be computed putting more weight on the gaze.  

\textbf{Approach $3$.} Instead of using domain knowledge (approach 1) or manually weighing the modalities (approach 2), the expressiveness score $\tau$ can be estimated using a linear generative model with Gaussian latent variables \cite{bishop2006pattern}. Here, we feed all modalities to the generative model as factors to compute latent facial expressiveness variable ($\tau_{fa}$). 

\begin{figure}
\centering
\begin{subfigure}{.15\textwidth}
  \centering
  \includegraphics[width=.99\linewidth]{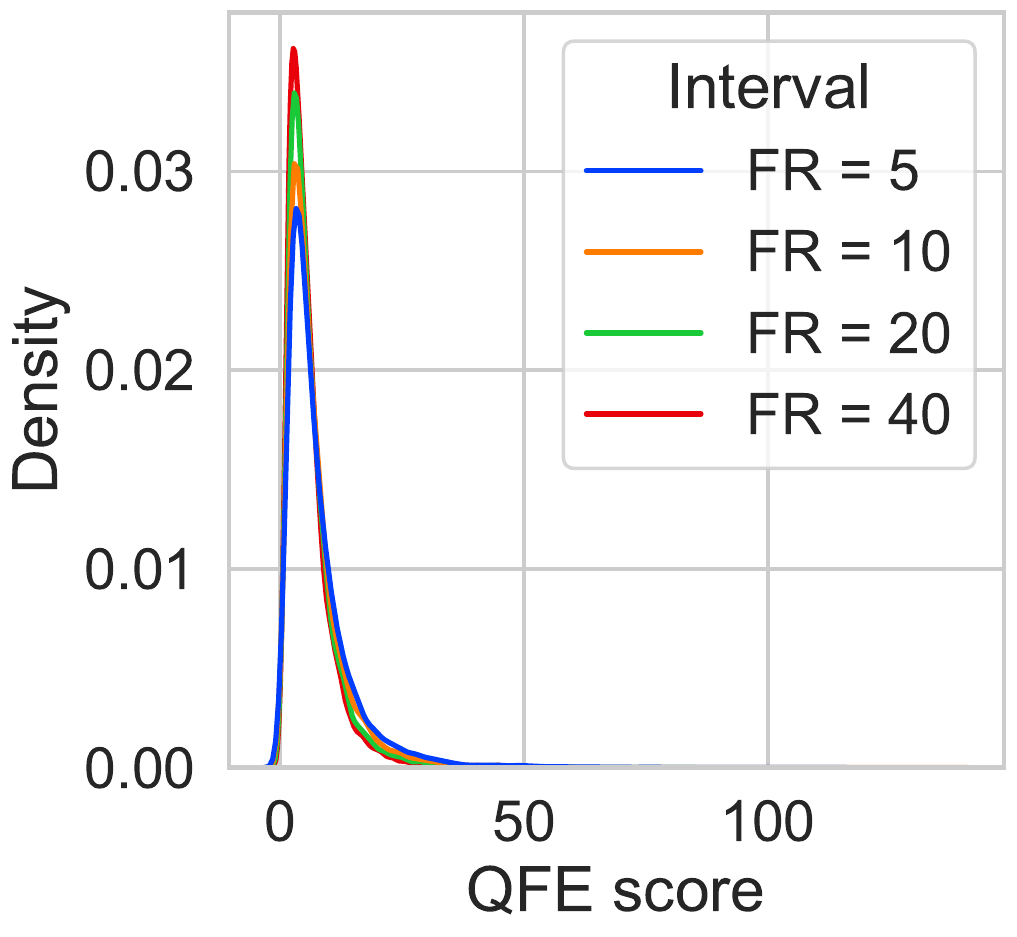}
  \caption{Approach 1}
  \label{fig:qfe_phy_df}
\end{subfigure}%
\begin{subfigure}{.15\textwidth}
  \centering
  \includegraphics[width=.99\linewidth]{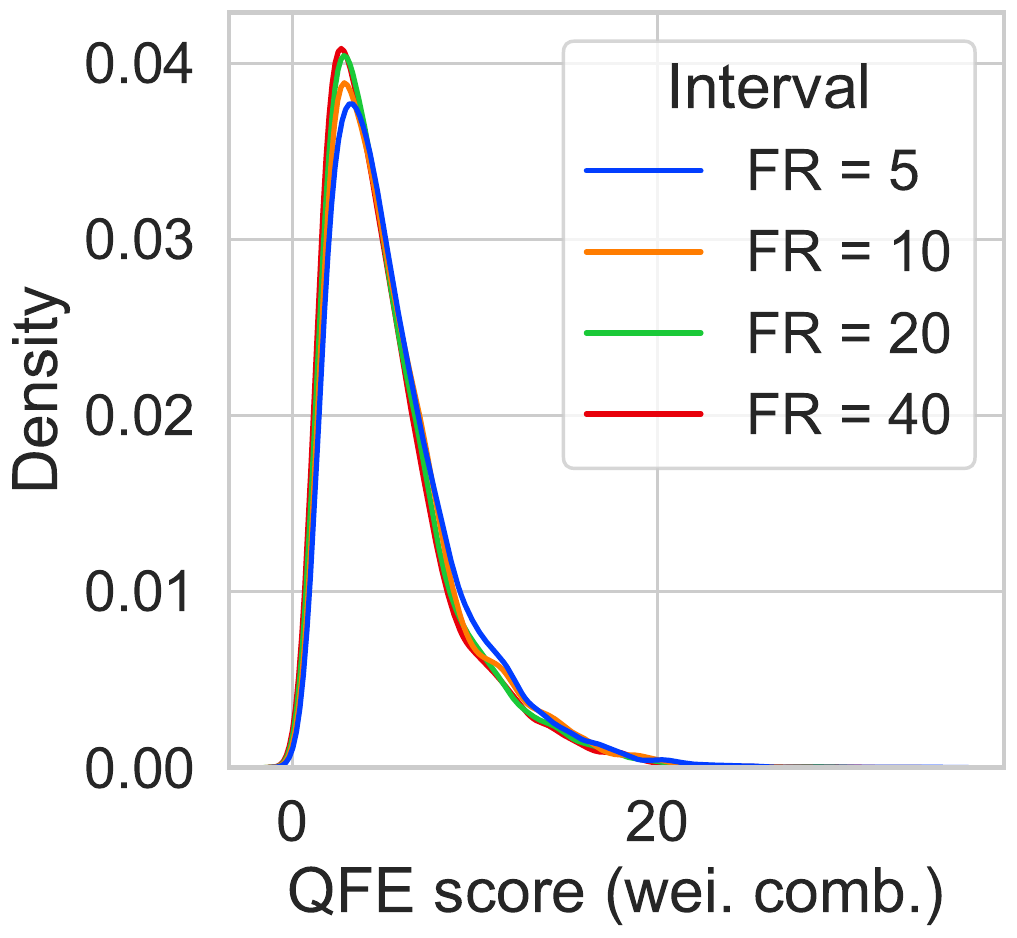}
  \caption{Approach 2 }
  \label{fig:qfe_scr_linComb}
\end{subfigure} 
\begin{subfigure}{.15\textwidth}
  \centering
  \includegraphics[width=.99\linewidth]{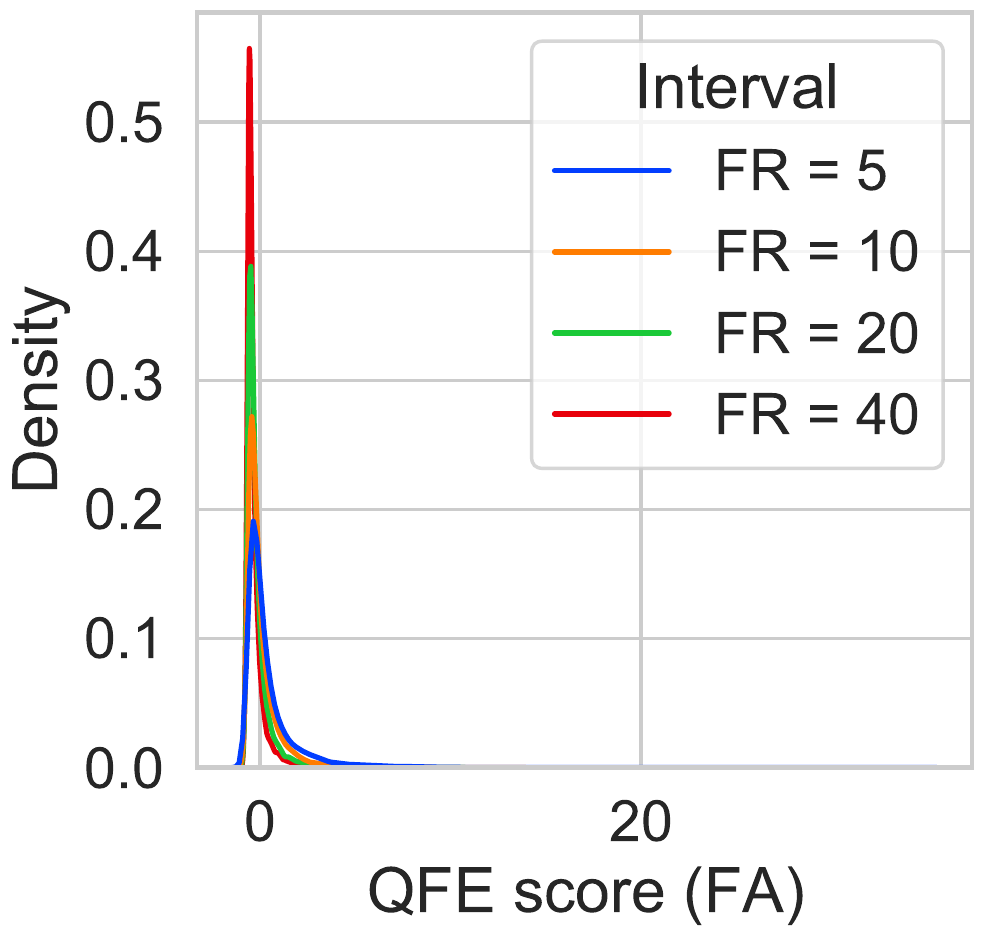}
  \caption{Approach 3 }
  \label{fig:qfe_scr_FA}
\end{subfigure} 
\vspace{-3mm}
\caption{QFE score distribution from the 3 approaches.}
\label{fig:qfe_final_scr_all}
\vspace{-5mm}
\end{figure}

We refer the reader to Fig. \ref{fig:qfe_final_scr_all} for the QFE score distribution across these 3 approaches.

\vspace{-3mm}
\subsection{Affective Behavior Analytics} 
\vspace{-2mm}
The quantification of facial expressiveness is essential as $\tau$ provides detailed information captured from both the spatial and temporal expressiveness. Using $\tau$ can help perform affective science and emotion AI research at scale, given that it has the potential to enhance data collection and hypothesis testing on large-scale datasets, while incorporating context. This has the potential to augment affective behavior analytics. To demonstrate use cases of the QFE in emotion research and affective computer vision, in this section, we describe two important affective computing tasks.  


\textbf{Granger Causality between Temporal Context and Affective Facial Expressions}. In emotion research and affective computer vision, stimuli or context are used to elicit facial expressions on subjects. One natural question that arises is that for a given stimulus, \textit{can we measure whether the stimulus elicited the facial expression?} In this work, we use the QFE score $\tau$ and temporal context to test the relationship between stimuli and facial expressions. We formulate the problem as follows: assuming $\tau$ and the stimulus are temporal variables, we use the Granger causality \cite{seth2007granger} test to evaluate whether stimulus Granger-causes the expressiveness. We represent this as $GC(c) \longrightarrow \tau$
where $c$, and $GC$ are the stimulus, and Granger causality, respectively. We can say $c$ Granger-causes $\tau$ if the historic values of stimulus $c$ can predict the future value of the $\tau$. If we find significant evidence of $c$ Granger-causing $\tau$, then we can conclude that stimulus is able to elicit facial expressiveness. This along with the proposed $\tau$ has the potential to explore the relationship between context and expressiveness at scale to augment emotion research, and evaluate users' responses to multimedia content. See Sec. \ref{sec:exp} for more details. 


\textbf{Quantifying Subjectivity in Context}. Affect is highly subjective \cite{kochan2013subjectivity, ledoux2018subjective, barrett2017emotions} due to factors including, but not limited to, personality, gender, and culture. To develop automated affect perception models, a sound understanding and quantitative analysis of subjectivity of facial expressiveness is required. In this work, we demonstrate how quantified facial expressiveness can be exploited to quantify the subjective difference among people. To do so, we first compute the QFE score $\tau$ for each subject in a given context, and then, we perform several statistical measurements to quantify the difference. See Sec. \ref{sec:exp} for more details.

\begin{figure}
\centering
\begin{subfigure}{0.49\textwidth}
  \centering
  \includegraphics[width=.99\linewidth]{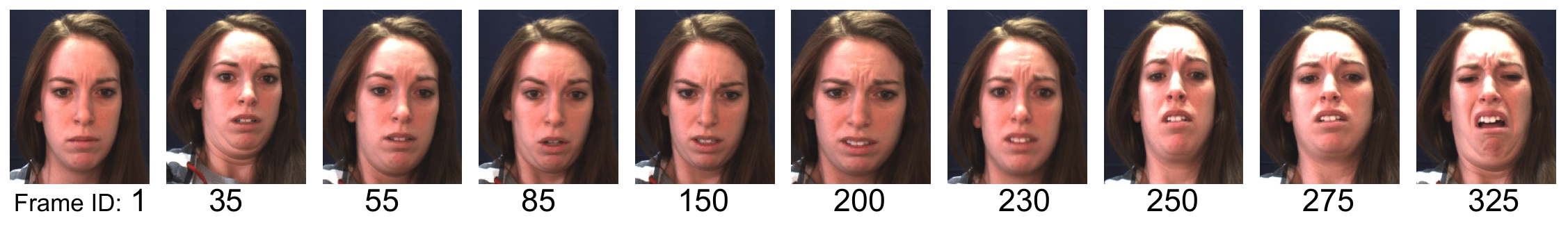}
  \label{fig:uaeALgo_ex1}
\end{subfigure} 
\begin{subfigure}{0.49\textwidth}
  \centering
  \includegraphics[width=.99\linewidth]{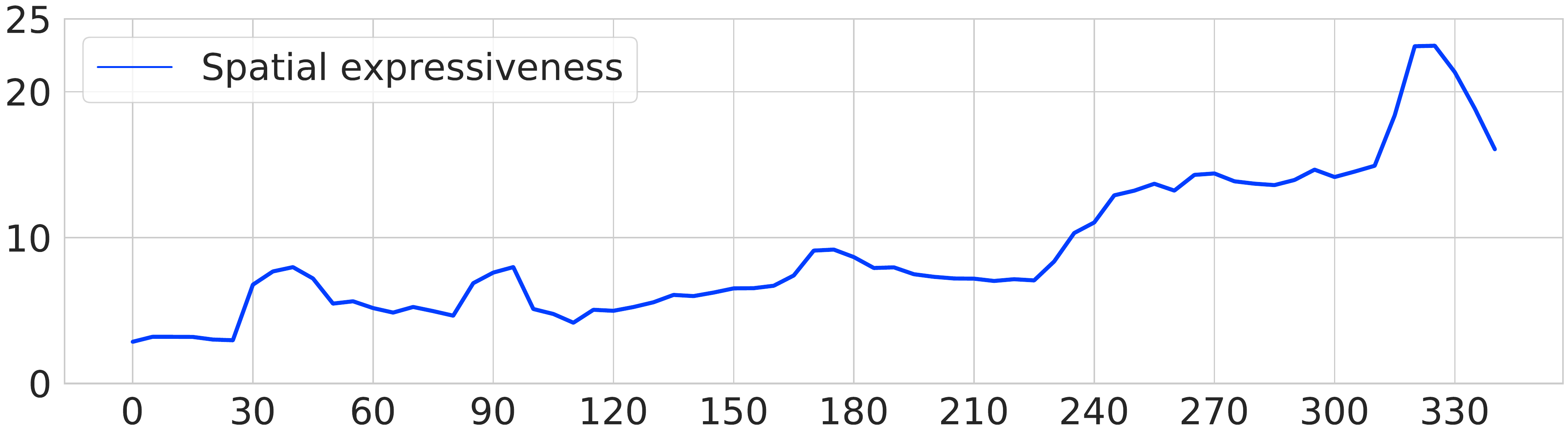}
\end{subfigure}\par
\begin{subfigure}{0.49\textwidth}
  \centering
  \includegraphics[width=.99\linewidth]{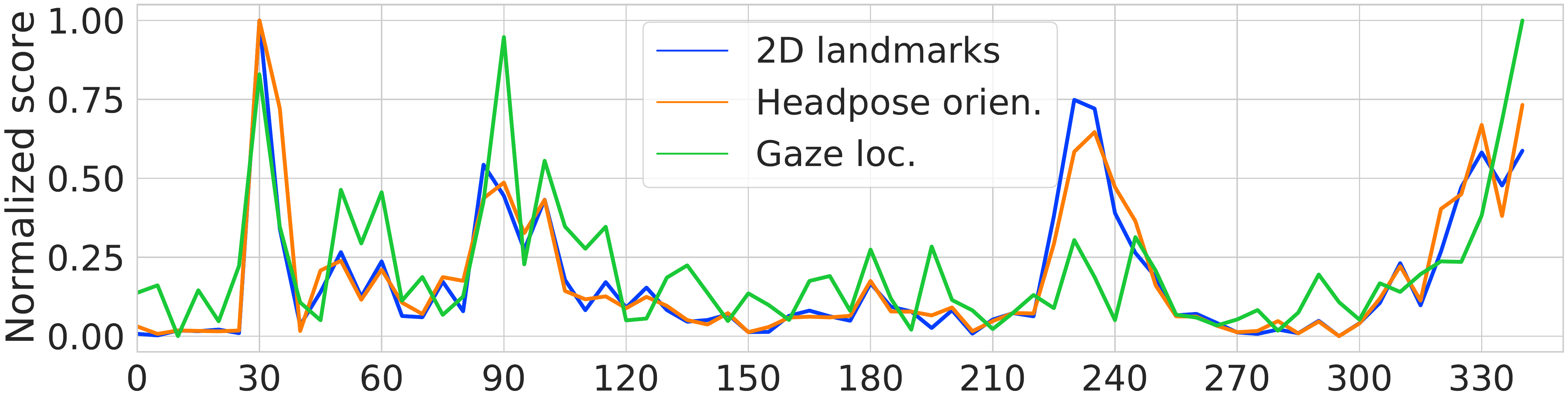}
\end{subfigure} 
\begin{subfigure}{0.49\textwidth}
  \centering
  \includegraphics[width=.99\linewidth]{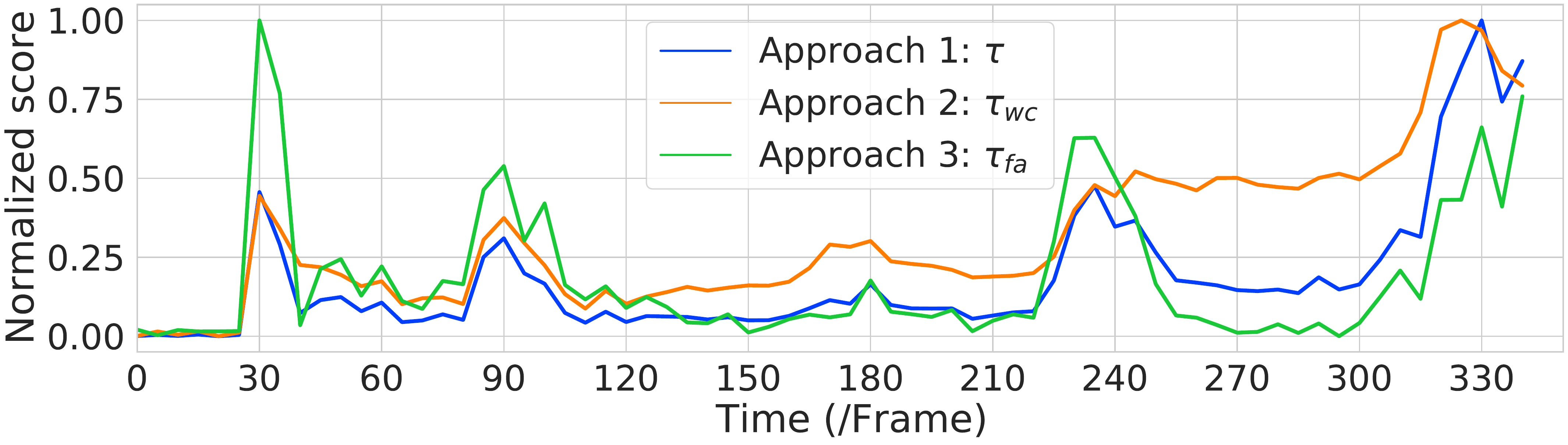}
  \label{fig:qfe_final_scr_SN011}
\end{subfigure} 
\vspace{-3mm}
\caption{Sample sequence with computed quantified expressiveness scores. Top to bottom: frames from the sequence, the magnitude of the spatial expressiveness $\sigma$, temporal expressiveness: 2D landmarks, headpose orientation, gaze location and QFE scores: $\tau, \tau_{wc}, $ and $\tau_{fa}$. For visualization purposes, temporal expressiveness, and $\tau, \tau_{wc}, $ and $\tau_{fa}$ are normalized in between $[0, 1]$. (Best viewed in color).}
\label{fig:uaeAlgoEx1}
\vspace{-5mm}
\end{figure}
\vspace{-4mm}
\section{Experiments and Analysis}
\label{sec:exp}
\vspace{-3mm}

\begin{figure*}
\centering
\begin{subfigure}{.245\textwidth}
  \centering
  \includegraphics[width=.99\linewidth]{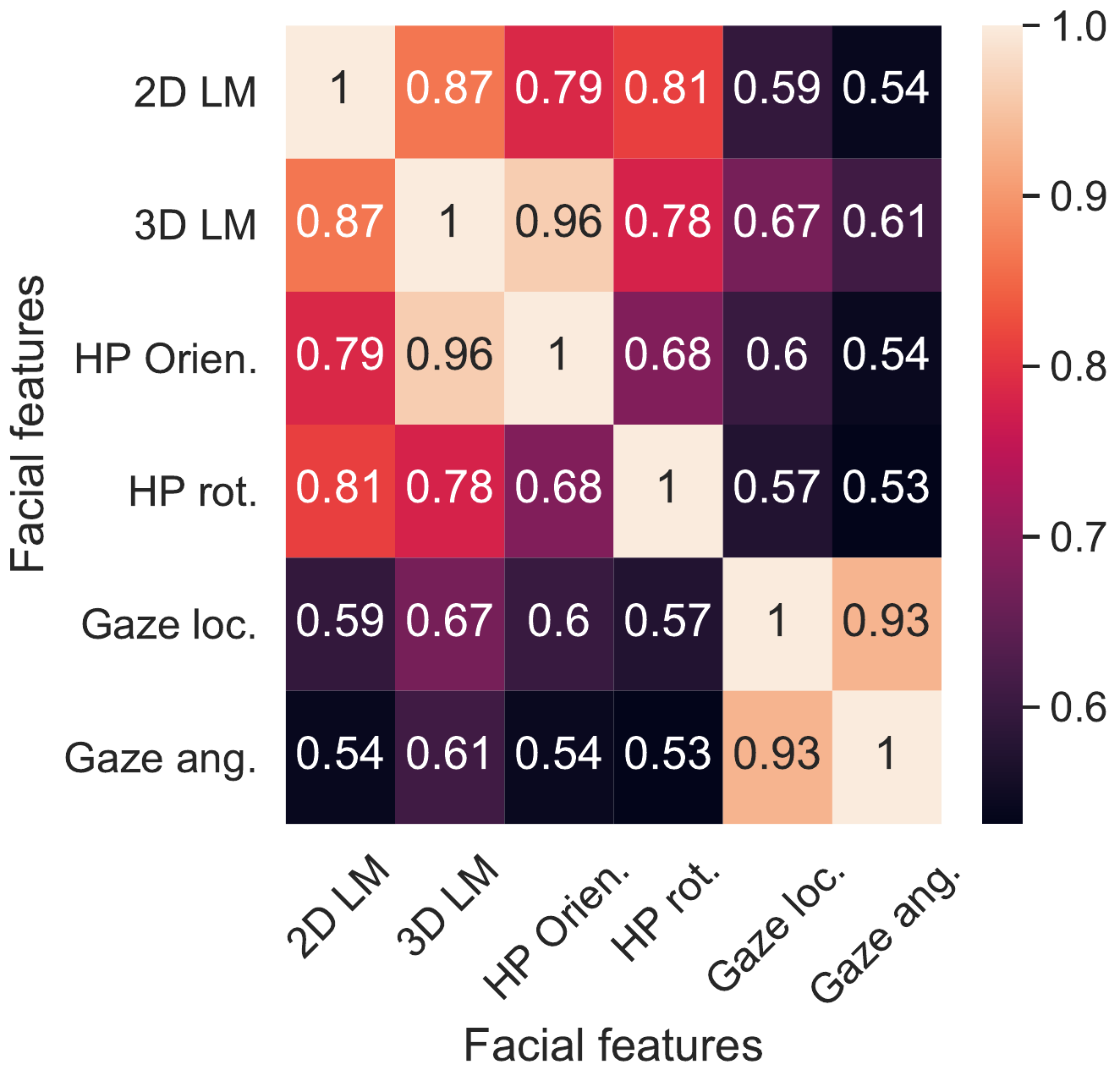}
  \caption{FR = 5}
  \label{fig:sampSeq1}
\end{subfigure}%
\begin{subfigure}{.245\textwidth}
  \centering
  \includegraphics[width=.99\linewidth]{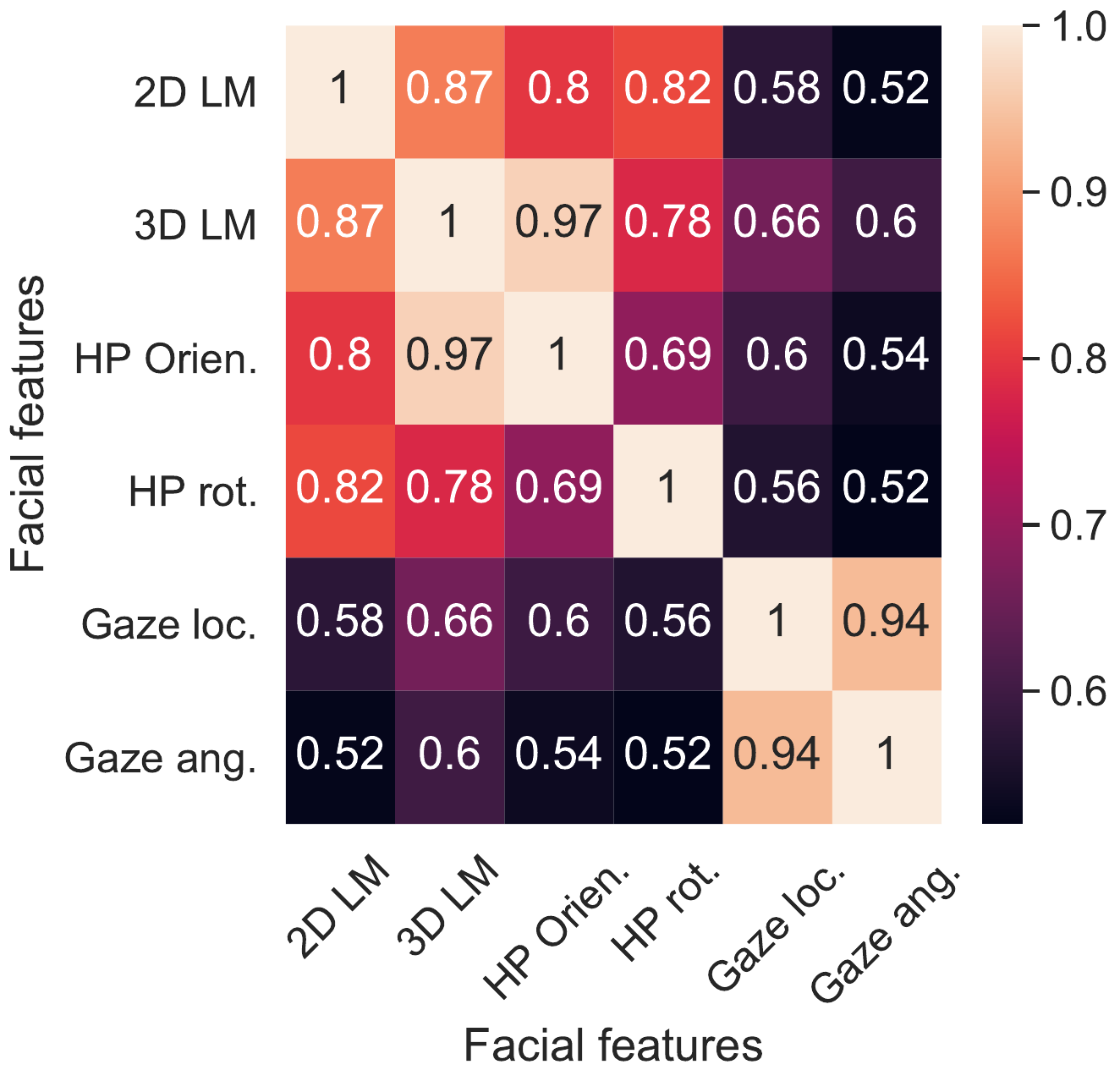}
  \caption{FR = 10}
  \label{fig:sampSeq2}
\end{subfigure} 
\begin{subfigure}{.245\textwidth}
  \centering
  \includegraphics[width=.99\linewidth]{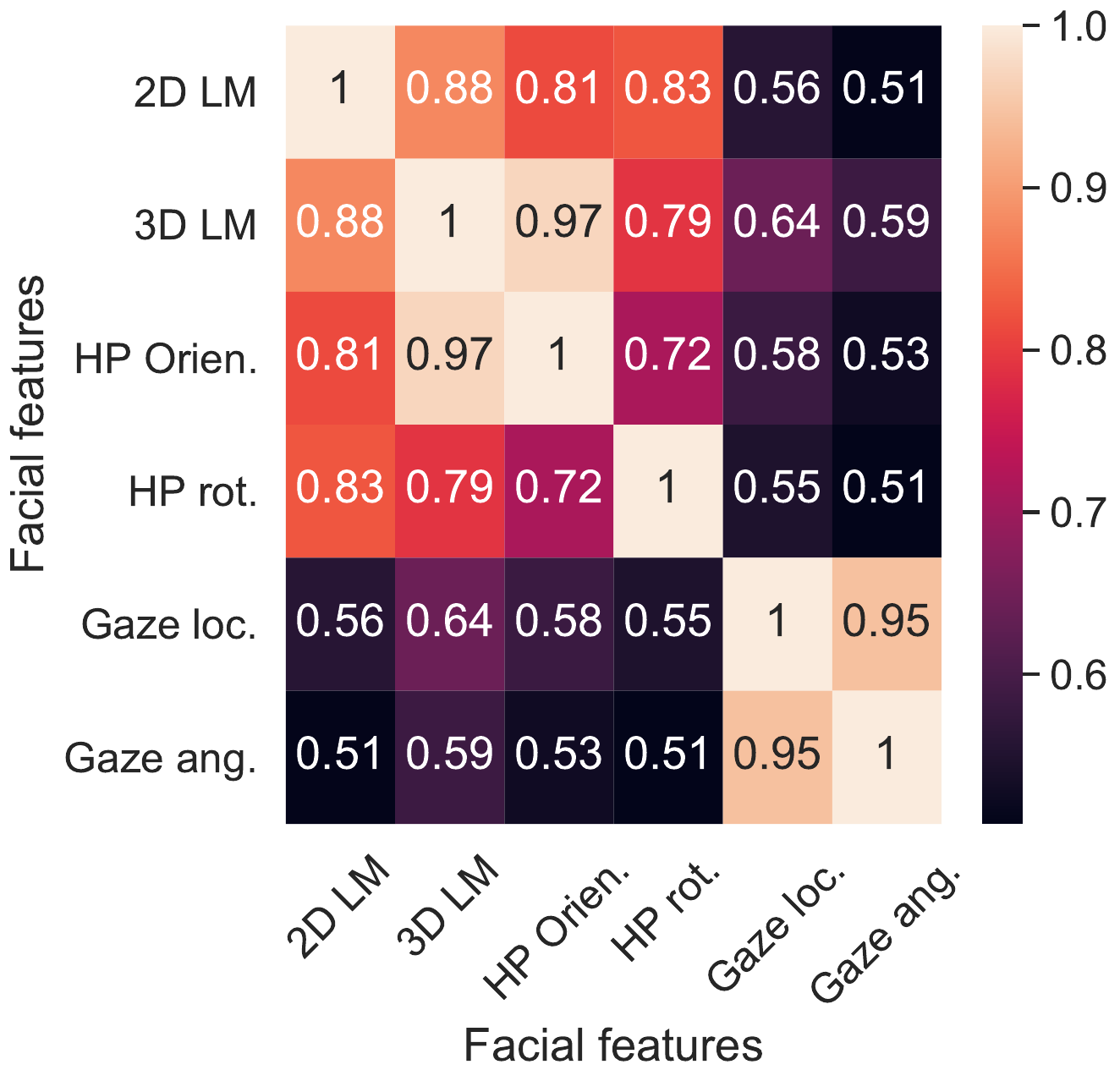}
  \caption{FR = 20}
  \label{fig:sampSeq3}
\end{subfigure} 
\begin{subfigure}{.245\textwidth}
  \centering
  \includegraphics[width=.99\linewidth]{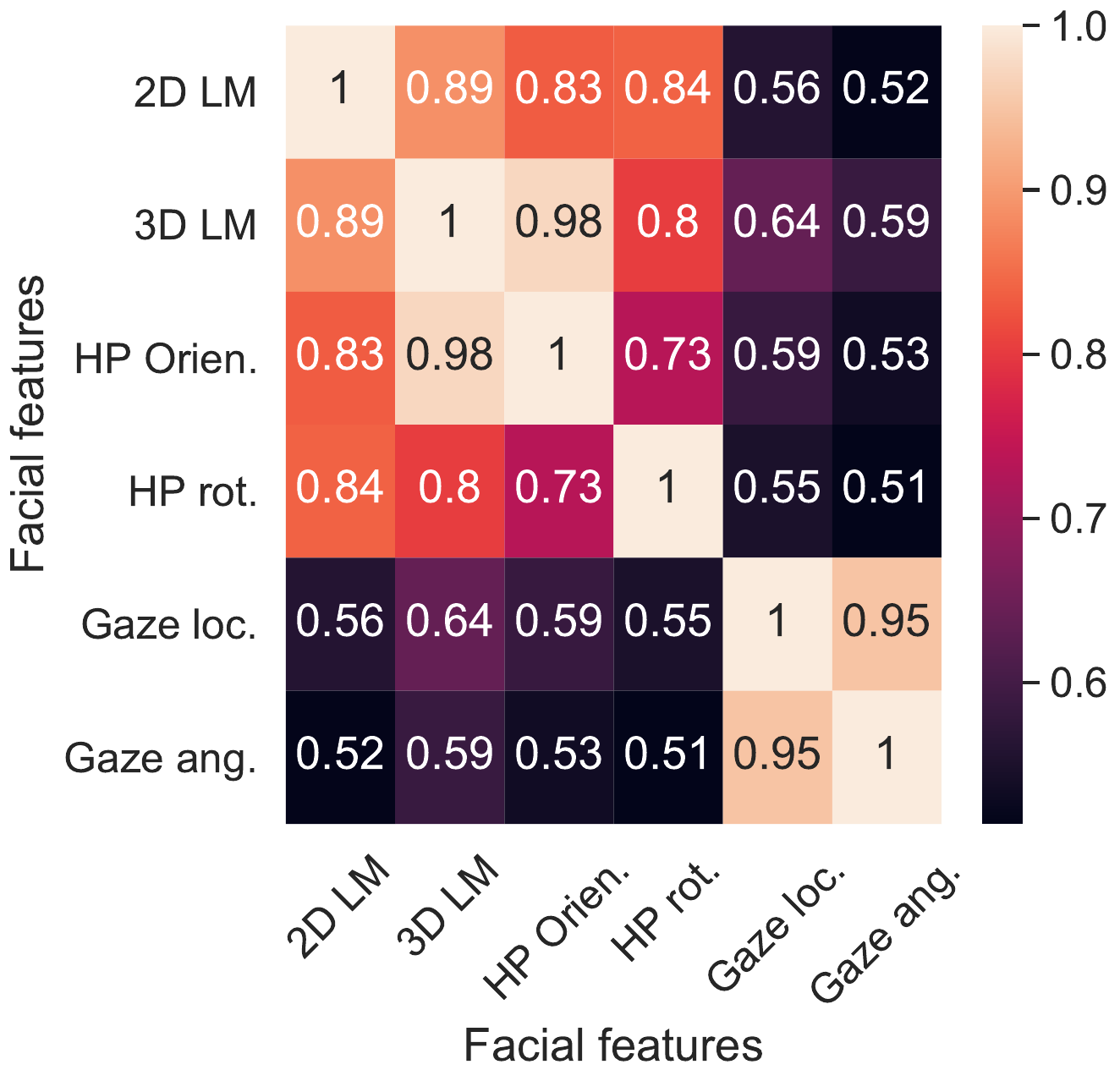}
  \caption{FR = 40}
  \label{fig:sampSeq3}
\end{subfigure} 
\vspace{-3mm}
\caption{Association among candidate temporal facial features. (Best viewed in color and zoomed in).  } 
\label{fig:corrComb_dynmFeat}
\vspace{-3mm}
\end{figure*}

\begin{figure*}
\centering
\begin{subfigure}{.2\textwidth}
  \centering
  \includegraphics[width=.99\linewidth]{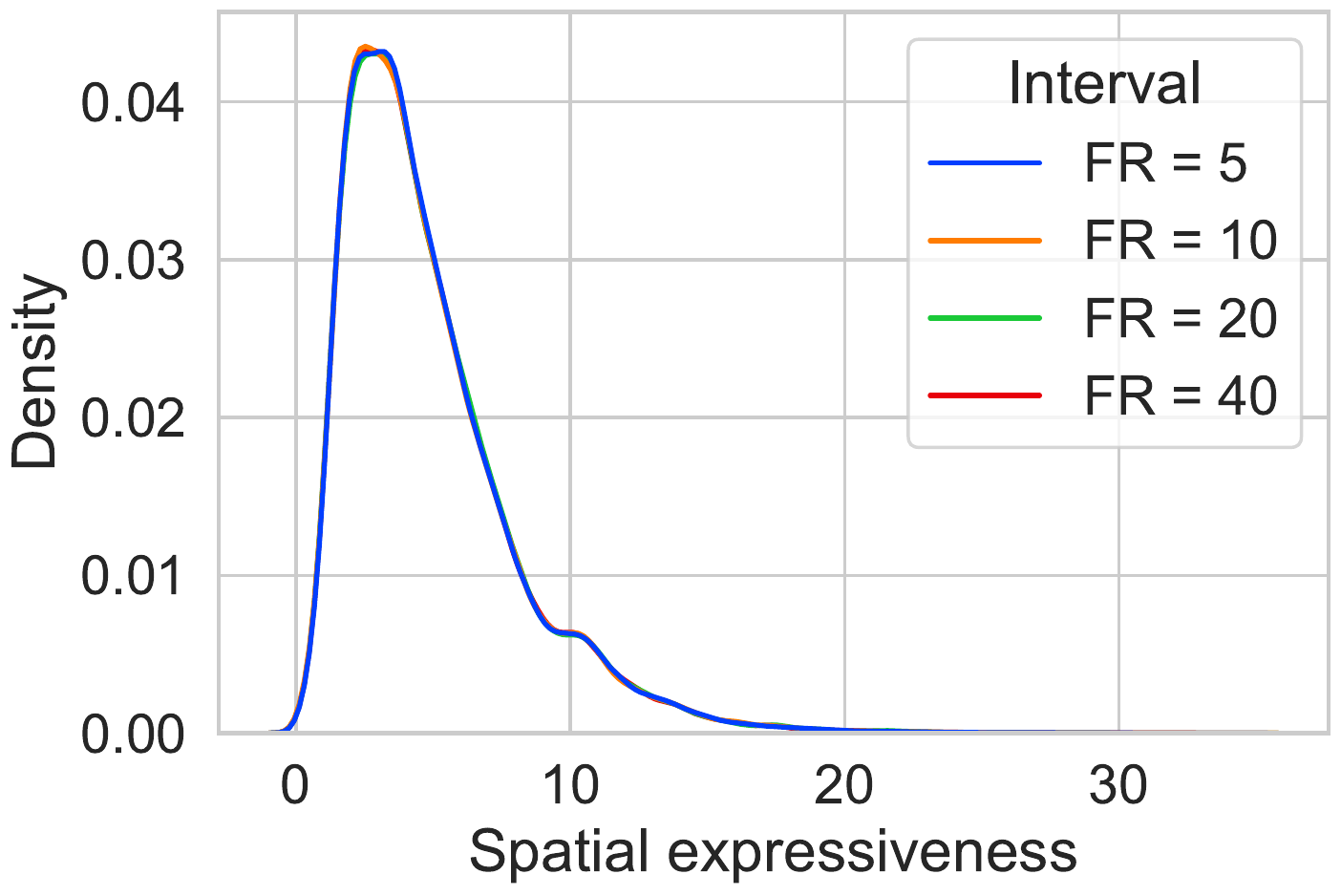}
  \caption{$\sigma$}
  \label{fig:KDE_Spatial_expressiveness}
\end{subfigure}%
\begin{subfigure}{.2\textwidth}
  \centering
  \includegraphics[width=.99\linewidth]{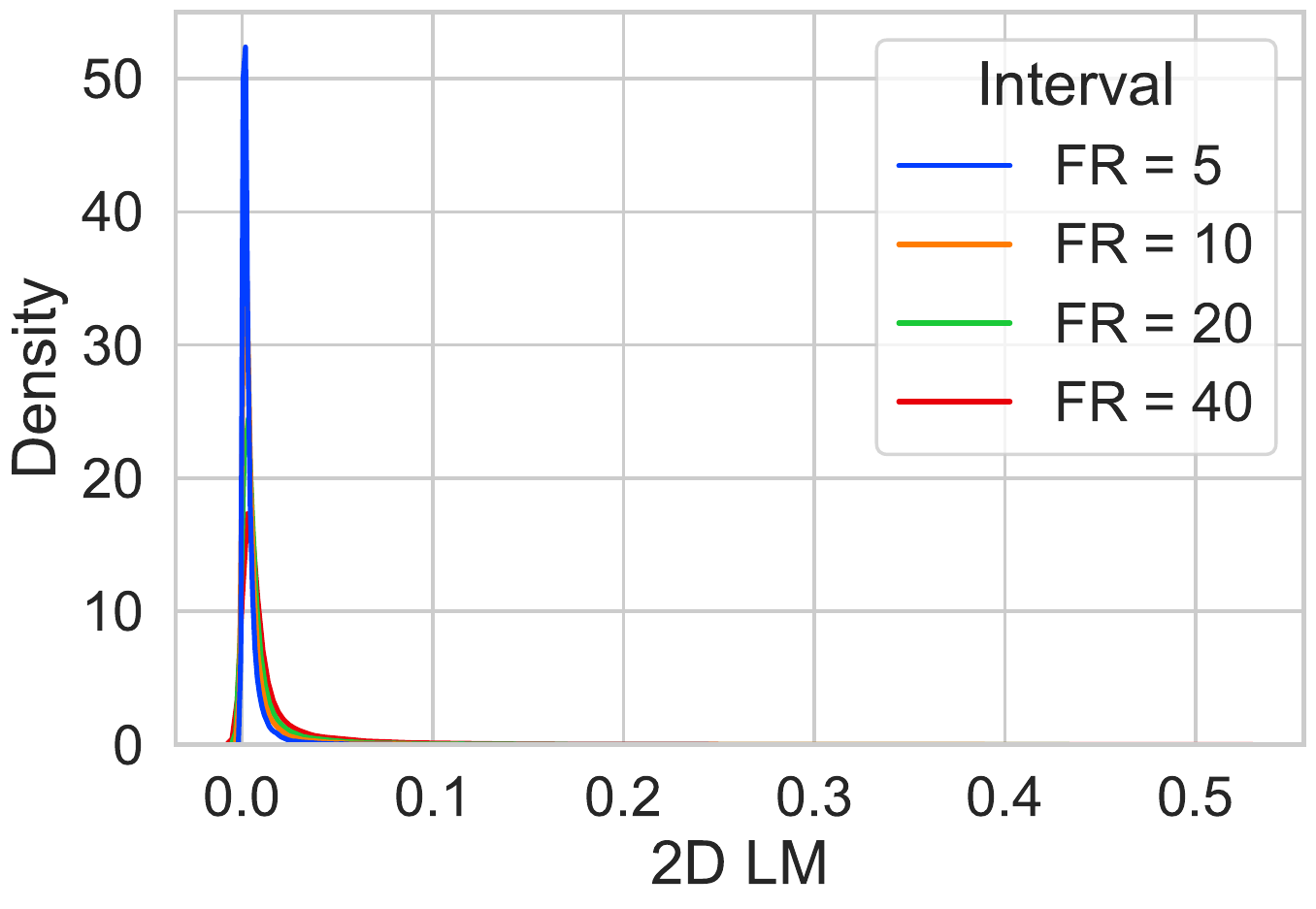}
  \caption{$\delta_{lm}$}
  \label{fig:KDE_2D_LM}
\end{subfigure} 
\begin{subfigure}{.2\textwidth}
  \centering
  \includegraphics[width=.99\linewidth]{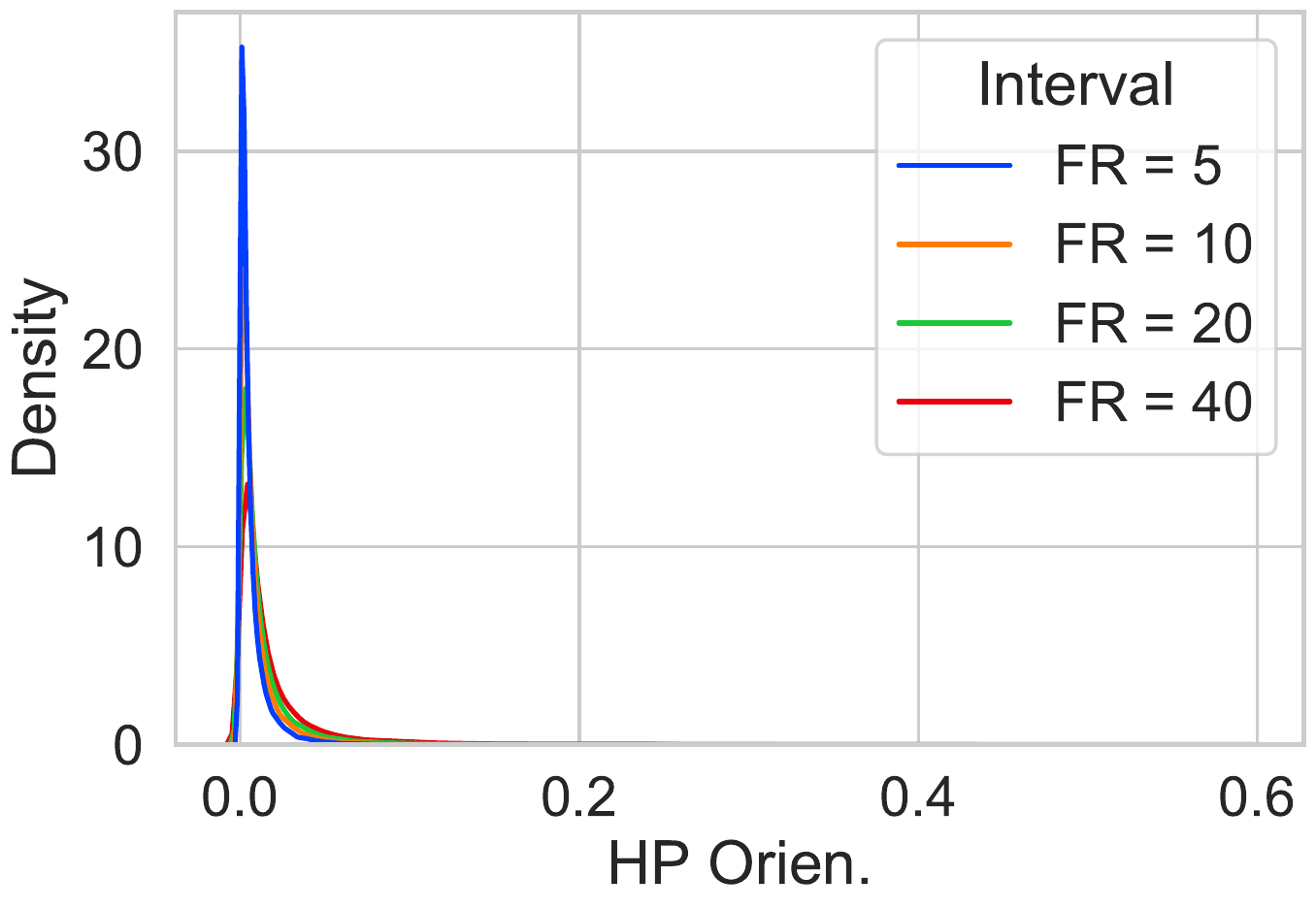}
  \caption{$\delta_{hp}$}
  \label{fig:KDE_HP_Trans}
\end{subfigure} 
\begin{subfigure}{.2\textwidth}
  \centering
  \includegraphics[width=.99\linewidth]{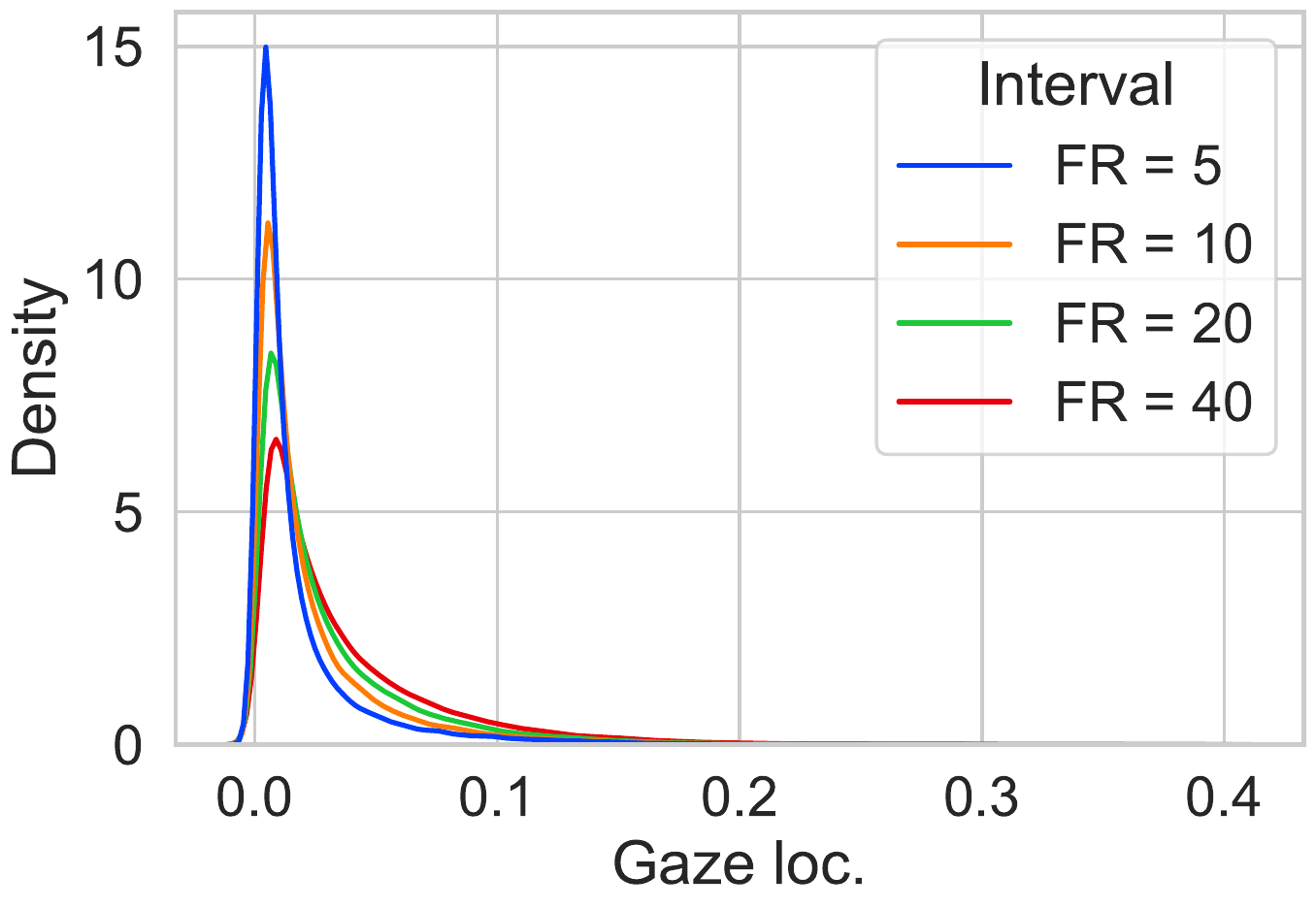}
  \caption{$\delta_{g}$}
  \label{fig:KDE_Gaze_loc}
\end{subfigure} 
\vspace{-3mm}
\caption{Spatial and temporal expressiveness distribution (Best viewed in color and zoomed in).}
\label{fig:corrDynmFeatDist}
\vspace{-5mm}
\end{figure*}

As the goal of this work is to quantify facial expressiveness in a given moment in time and demonstrate use cases, we first computed the QFE score using the DISFA and BioVid pain datasets. Then, we experimented with two downstream tasks, namely Granger causality analysis among modalities (e.g. context, QFE score, ground truth), and subjectivity quantification. These are two important tasks that are essential in emotion research and applied affective compter vision given the constructed theory on emotion and it's relationship to context \cite{siegel2018emotion, barrett2019emotional} and face as sensing, as well as human subjectivity \cite{kochan2013subjectivity, ledoux2018subjective}.

\textbf{Data preparation}. To track and extract the facial features such as landmarks (LM), headpose (HP), eye gaze (G), and facial AU intensities, we used OpenFace \cite{baltrusaitis2018openface}, which is a publicly available facial behavior analysis tool. Since $LM, HP,$ and $G$ represent spatial and depth information of the face and do not have expression intensity in the same way we have for AUs, we normalized $LM, HP, $ and $G$ in between $[0, 1]$ using min-max normalization \cite{han2011data}.

\vspace{-2mm}
\subsection{Datasets}
\vspace{-3mm}
\textbf{DISFA dataset} \cite{mavadati2013disfa} is a publicly available spontaneous facial expression dataset which contains $27$ subjects (12 females, and 15 males) aged in between $[18, 50]$ and ethnically $3$ Asian, $1$ Black, $21$ Caucasian, and $2$ Hispanic subjects. The dataset contains frontal face images and action unit (AU) \cite{ekman1976measuring} annotations at frame level, by expert annotators. The dataset contains $27$ videos comprising $130,000$ images. A video comprising of $9$ segments with different types of content was used as the stimuli to elicit the expression. The stimuli video and corresponding frontal face videos of the subjects are each $242$ seconds long. 

\textbf{BioVid pain dataset} \cite{walter2013biovid} contains $90$ subjects performing pain and other expressions. The dataset is balanced in terms of gender. There are three age groups in the dataset in the age range of $ [18-35], [36-50], [51-65]$ years old. In this work, we used the raw data which is part C in the data portion and contains $87$ subjects. Continuous heat (temperature) was used as a stimulus to elicit pain expression, which was self-calibrated by the subjects. The length of each session is approximately $25$ minutes, and in total, in Part C, is comprised of approximately $3.26$ million images. Aside from the frontal face videos and temperature, the dataset contains pain labels in between $[0, 4]$, where $0$ means no pain, and $4$ means maximum pain. 
\begin{table}
 \caption{Descriptive summary of facial expressiveness scores $\tau$ for given pain level on BioVid pain dataset. Here, $25\% P.$ and $75\% P.$ denote $25^{th}$ and $75^{th}$ percentiles. We can see that with the change in the pain level (PL), $\tau$ is not changing much. Especially PL $1, 2, $ and $3$, which are very similar in terms of expressiveness. This questions the efficacy of heat as the stimuli to elicit facial pain expressions. The $\tau$ summary also partially explain the failure to pass the GC test in Table \ref{tab:GC_res}. This summary is also aligned with the findings of Werner et al. \cite{werner2017analysis} as they pointed out the weak and low facial pain response for PL 1 and 2.}
  \centering
  \begin{tabular}{L{0.3cm}|L{0.75cm}C{0.5cm}C{0.4cm}C{0.7cm}C{0.7cm}C{0.7cm}C{0.75cm}}
    \toprule
     PL & Mean & SD	& Min. & $25\%$ P. & Med.	& $75\%$ P. & Max.\\
    \midrule
    1 & 8.2 & 8.2 & 0 & 2.8 & 5.8 & 10.8 & 139.8 \\ 
    2 & 8.3 & 8.4 & 0 & 2.8 & 5.8 & 11.1 & 144 \\
    3 & 8.6 & 9.2 & 0 & 2.8 & 5.7 & 11.4 & 137.3 \\
    4 & 10.0 & 10.9 & 0 & 3.2 & 7.0 & 13.0 & 146.3 \\
    \bottomrule
  \end{tabular}
   \vspace{-5mm}
  \label{tab:exprv_scr_summ}
\end{table}
\subsection{Quantified Facial Expressiveness}
\label{sec:qfe}
\vspace{-2mm}
To compute the quantified facial expressiveness (QFE) score, we computed the spatial expressiveness $\sigma$ using Eqn. \ref{eqn:staticExprv}. Note that $\sigma$ can be computed using all available AUs, or a subset of AUs depending on the context and task. In this work, we computed the \textbf{overall} spatial (static) expressiveness of the human face for a given video frame using 
AUs: $1, 2, 4, 5, 6, 7, 9, \- 10, 12, 14, 15, 17, 20, 23, 25, 26, 28$, and $45$ on both datasets. We also computed spatial expressiveness in the context of \textbf{pain} using pain-related AUs \cite{werner2019automatic}: $4, 6, 9, 10, 25$ on BioVid. We set $\lambda=100$ to have $\sigma$ in between $[0, 100]$. We then computed the temporal expressiveness using landmarks, head pose, and eye gaze using Eqn. \ref{eqn:tempoExprv}, where $\infty = 20$. We found little change with $ \infty > 20$.

\begin{figure}
\centering
\begin{subfigure}{0.49\textwidth}
  \centering
  \includegraphics[width=.99\linewidth]{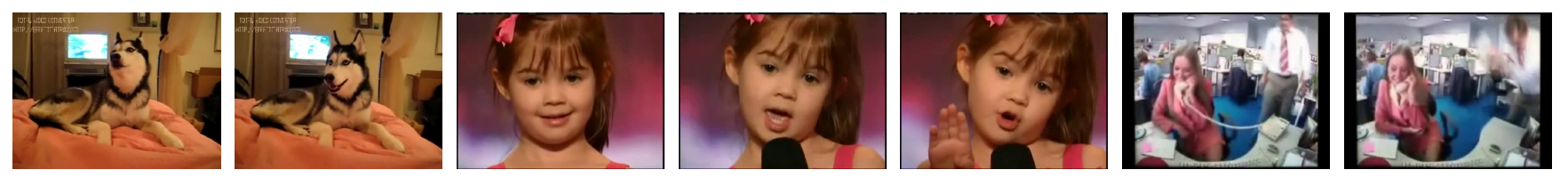}
  \label{fig:DISFA_stimuli_vid}
\end{subfigure}\par
\begin{subfigure}{0.49\textwidth}
  \centering
  \includegraphics[width=.99\linewidth]{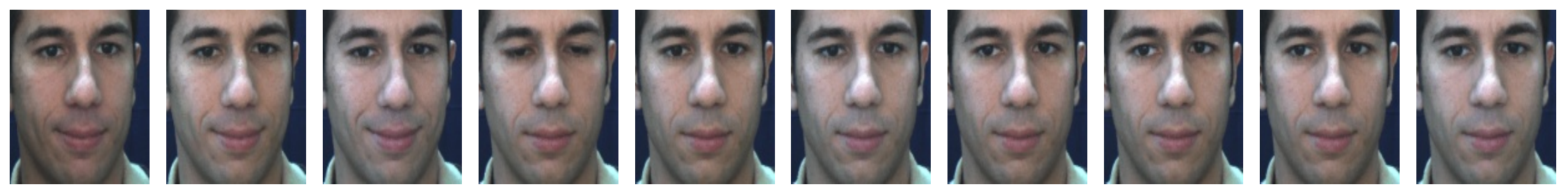}
  \label{fig:subDiff_SN001}
\end{subfigure}\par
\begin{subfigure}{0.49\textwidth}
  \centering
  \includegraphics[width=.99\linewidth]{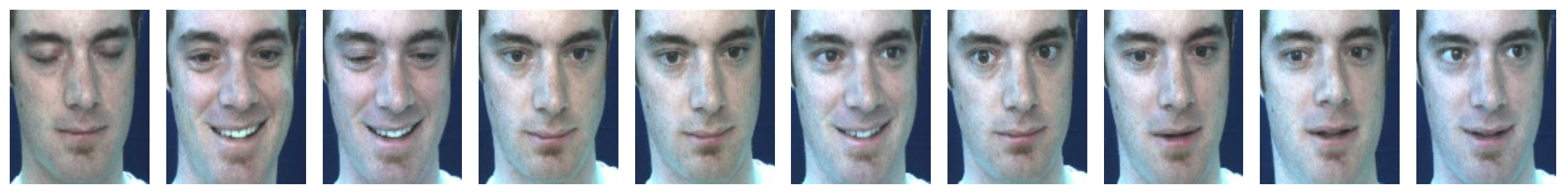}
  \label{fig:subDiff_SN002}
\end{subfigure}\par
\begin{subfigure}{0.49\textwidth}
  \centering
  \includegraphics[width=.99\linewidth]{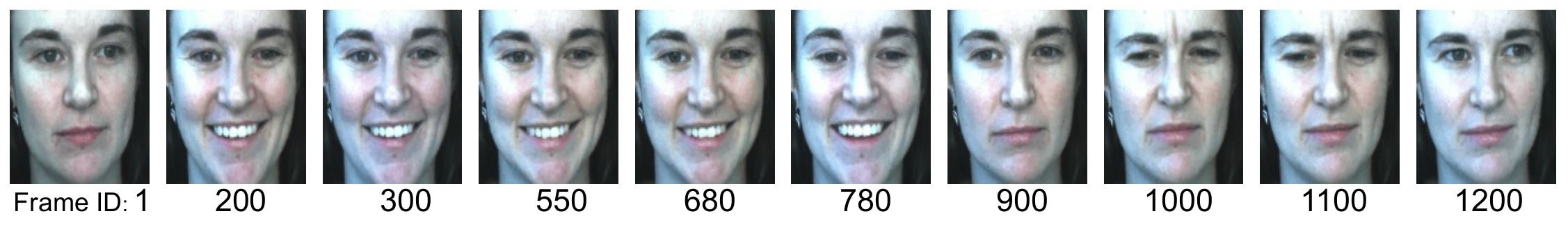}
  \label{fig:subDiff_SN003}
\end{subfigure} 
\begin{subfigure}{0.49\textwidth}
  \centering
  \includegraphics[width=.99\linewidth, height=25mm]{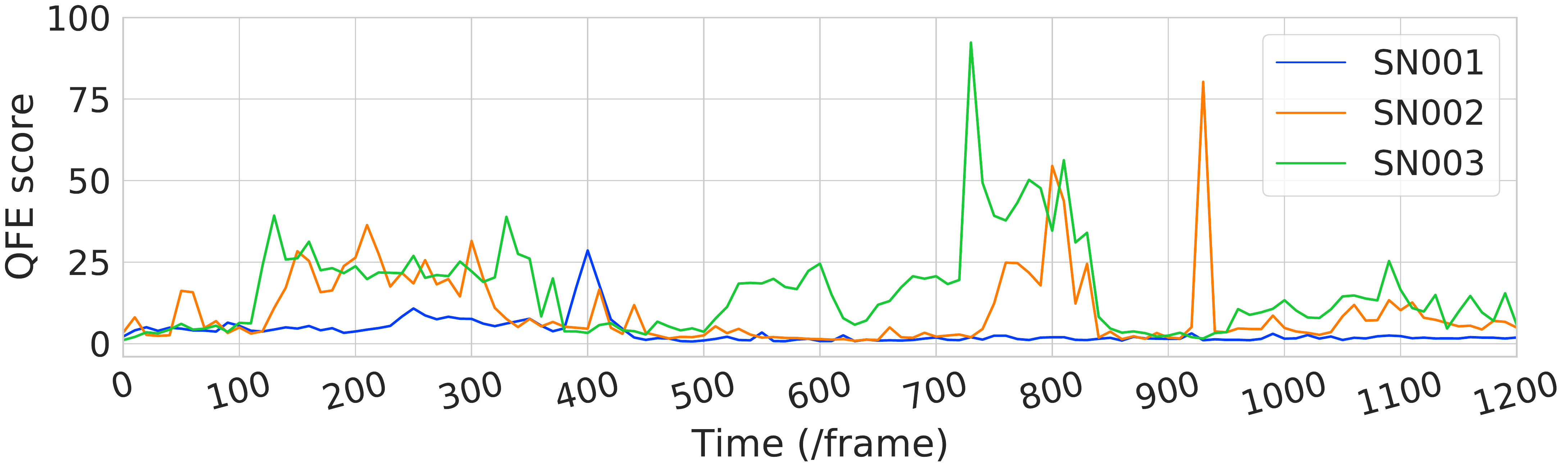}
  \label{fig:qfe_subDiff_example}
\end{subfigure} 
\vspace{-3mm}
\caption{Example video segments from DISFA showing subjective differences with same context. Top to bottom: peak frames from stimulus video, frames from subjects SN001, SN002, SN003, and QFE scores $\tau$ computed for each subject, respectively. (Best viewed in color). }
\label{fig:sampSeqSubDiff}
\vspace{-7mm}
\end{figure}
\vspace{-1mm}
\textbf{Ablation study on frame rate, and temporal modalities (features)}. This ablation study is performed by sampling data from both DISFA and BioVid pain datasets. We performed an ablation study on impact of frame rate (FR) (interval at which we picked two frames to compute the difference (i.e. $\Delta_x$ in Eqn. \ref{eqn:vel})), and facial features used to capture the temporal expressiveness. More precisely, we experimented with $2D$ and $3D$ landmakrs (\textbf{LM}), orientation and rotation of headpose (\textbf{HP}), and location and angle of eye gaze (\textbf{G}), while setting $FR$ to $5, 10, 20,$ and $40$, respectively. We measured the association among the modalities using Spearman rank correlation coefficient (\textbf{SRCC}) \cite{benesty2009pearson}, and the obtained results are reported in Fig. \ref{fig:corrComb_dynmFeat}. We found that 2D LM, 3D LM, and HP orientation and rotation are moderately positive to strongly correlated. We also found that gaze location and angle are strongly correlated. To keep a balance between modalities and to reduce the redundant computation, we selected $2D$ LM, HP orientation, and gaze location to capture the temporal expressiveness in our algorithm. Fig. \ref{fig:corrDynmFeatDist} depicts the distribution (Estimated using kernel density estimation (KDE) method) of $\sigma$, 2D LM, HP orientation, and G location setting FR to $5, 10, 20$, and $40$ in which we can observe that aside from the distribution of gaze, the shape of distribution did not change much with the change in FR. Hence, in our downstream tasks, we set FR to $5$ to compute the temporal expressiveness.

The expressiveness score $\tau$ is computed using the three approaches. In approach $1$, we computed $\tau$ assuming $\sigma$ is the major source of expressiveness, and $\delta$ as the minor source of expressiveness; hence, we set $\lambda_k$ in Eqn. \ref{eqn:QFE_scr} to $100, 100$, and $50$ for computing $\delta$ using 2D LM ($\delta_{lm}$), using HP orientation ($\delta_{hp}$), and using G location ($\delta_{g}$), respectively. In approach $2$, we computed the weighted combination of $\sigma$, $\delta_{lm}, \delta_{hp}, $ and $\delta_{g}$ in which we set each weight $w = 100$, and $\epsilon = 0$. Finally, in approach $3$, we used a latent variable model to estimate the $\tau$ feeding all available facial features. In case of approach $3$, we performed Bartlett’s test to evaluate the factorability of the input facial features, and Kaiser-Meyer-Olkin (KMO) test to evaluate suitability of data for factor analysis on the input features. The data passed the factorability test with $p-value < 0.001$, mean (standard deviation (\textbf{SD})) $KMO=0.727 \pm 0.02$, and $p-value < 0.001$, mean (SD) $KMO = 0.744 \pm 0.02$ across four FR settings for DISFA dataset and BioVid pain dataset, respectively. An example on computed expressiveness score is shown in Fig. \ref{fig:uaeAlgoEx1} in which we observe that the proposed method can capture the facial expressiveness, as all three approaches were able to measure the overall expressiveness of the face. In our next set of experiments, we used the $\tau$ computed using \textbf{approach $1$} since we focus on affective experience/responses for those tasks.
\begin{figure}
\centering
\begin{subfigure}{0.5\textwidth}
  \centering
  \includegraphics[width=.99\textwidth]{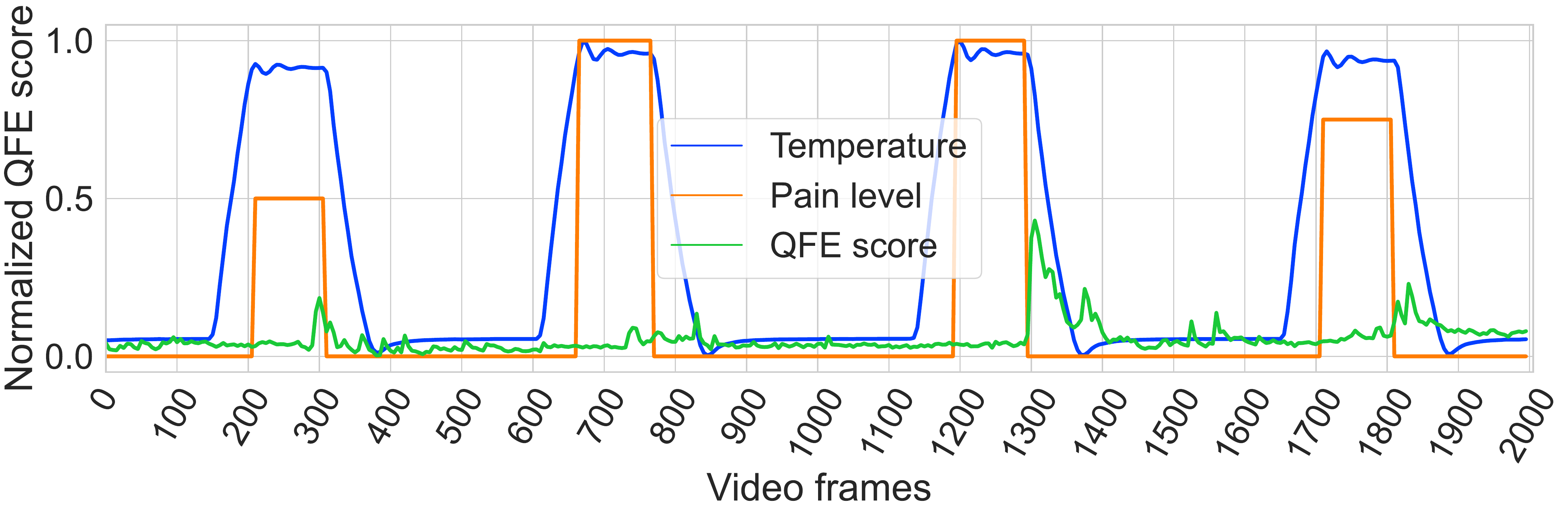}
\end{subfigure} 
\vspace{-3mm}
\caption{Association among temperature, QFE score ($\tau_{pain}$), and pain level (BioVid). For visualization purposes, temperature, $\tau_{pain}$, and pain level are normalized to $[0, 1]$. We excluded face images from the visualization following data usage policy of bioVid pain dataset \cite{walter2013biovid}. }
\label{fig:GC_testngEx}
\vspace{-5mm}
\end{figure}

\begin{figure}
\centering
  \centering
  \includegraphics[width=.75\linewidth]{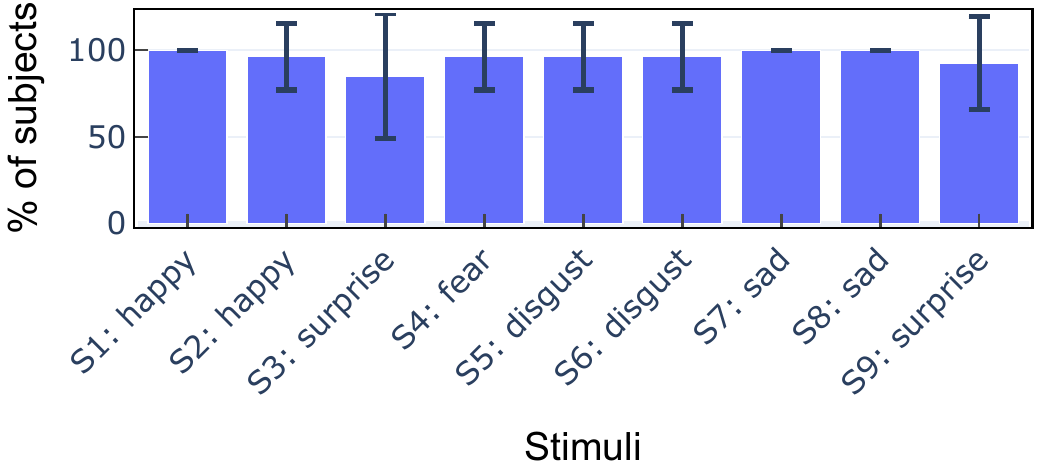}
  \vspace{-4mm}
  \caption{Percentage of subjects that responded, in terms of facial expressivity, to each stimulus in DISFA. Here, human annotators observed both stimuli video and QFE scores in parallel to identify whether a given stimulus video segment caused facial expressiveness on subject.}
  
\label{fig:DISFA_subResponded_to_stimuli}
\vspace{-7mm}
\end{figure}

\textbf{Evaluation via human annotators}: To measure the correctness of the QFE algorithm, we collected ratings from three annotators (2 males, 1 female), that were given instructions on rating before-hand. We used a questionnaire with three questions. $Q1$: 'Did the algorithm capture the expressivity? (response: yes/no)'; $Q2$: Rate the expressiveness score computed by the algorithm in between $[1, 5]$, where $1 =$ poor, $2 =$ weak, $3 =$ marginal, $4 =$ very good, $5 =$ excellent. We also asked the raters to provide their confidence on assessment ($Q3$), in between $[0, 100]$, (uncertain to certain).
We collected ratings for DISFA and BioVid datasets. In the case of DISFA, the entire sequence ($242$ seconds) was rated. In the case of BioVid, we selected 200 random sample ($5$ seconds long) sequences so that we can observe how QFE performed for both short and long sequences. The obtained rating summary is highlighted in T\-a\-ble \ref{tab:QFE_human_rating}. It can also be seen in Fig. \ref{fig:DISFA_subResponded_to_stimuli} that most subjects responded to the stimuli, however, some subjects did not responded to 'surprise', 'disgust', and 'fear'. Finally, \textit{even though most subjects responded to stimuli, the level and duration of responsiveness (facial expressiveness $\tau$) was variable and diverse as demonstrated in Sec. \ref{sec:sub_quant}.} These results are encouraging, as they show that QFE algorithm captured the expressiveness since the average assessments for both datasets falling in between 'very good' and 'excellent'.

\begin{table}
 \caption{QFE Algorithm evaluation via human annotators. 
 }
 \vspace{-2mm}
  \centering
  \begin{tabular}{L{1.5cm}|C{1.9cm}C{1.7cm}|L{1.5cm}}
    \toprule
     Datasets & Q1 & Q2	& Confidence \\
    \midrule
    DISFA & 1.0 & 4.6$\pm$ 0.46 & 96.7 $\pm$ 4.1  \\ 
    BioVid & 0.995 $\pm$ 0.07 & 4.58 $\pm$ 0.58 & 96.4 $\pm$ 4.1 \\ 
    \bottomrule
  \end{tabular}
  \label{tab:QFE_human_rating}
   \vspace{-4mm}
\end{table}

\begin{table*}
 \caption{Percentage (\%) of video segments for which temperature Granger-caused ground-truth pain level (PL), and facial pain expressiveness ($\tau_{pain}$). We set the significance level $\alpha$ to $0.05$. \textbf{PVSP} = percentage of video segment passed.
 }
 \vspace{-3mm}
  \centering
  \begin{tabular}{L{1cm}L{1cm}|C{1cm}C{1cm}C{1cm}C{1cm}C{1cm}|C{1cm}C{1cm}C{1cm}C{1cm}C{1cm}}
    \toprule
    
    \multicolumn{2}{c}{Lag} &  \multicolumn{5}{c}{\textbf{GC (temperature) $\longrightarrow$ PL}} & \multicolumn{5}{c}{\textbf{GC (temperature) $\longrightarrow$ $\tau_{pain}$ }} \\
     \cmidrule(r){1-2} \cmidrule(r){3-7} \cmidrule(r){8-12} 
     Time (sec.) & \# of frames &	PVSP LR test $\chi^{2}$	& PVSP params F test & PVSP SSR $\chi^{2}$ &	PVSP SSR F test F	& ALL & PVSP LR test $\chi^{2}$	& PVSP params F test & PVSP SSR $\chi^{2}$ &	PVSP SSR F test F	& ALL\\
    \midrule
    1 & 5 & 50.8 & 47.8 & 52.8 & 47.8 & 47.8 & 11.0 & 10.0 & 11.0 & 10.0 & 10.0 \\ 
    2 & 10 & 87.7 & 82.8 & 89.3 & 82.8 & 82.8 & 16.0 & 14.0 & 17.0 & 14.0 & 14.0 \\ 
    5 & 25 & 100.0 & 100.0 & 100.0 & 100.0 & 100.0 & 33.0 & 21.0 & 39.0 & 21.0 & 21.0 \\ 
    7 & 35 & 100.0 & 100.0 & 100.0 & 100.0 & 100.0 & 46.0 & 23.0 & 57.0 & 23.0 & 23.0 \\ 
    10 & 50 & 100.0 & 99.4 & 100.0 & 99.4 & 99.4 & 67.0 & 15.0 & 80.0 & 15.0 & 15.0 \\ 
    \bottomrule
  \end{tabular}
  \label{tab:GC_res}
  \vspace{-3mm}
\end{table*}

\textbf{Qualitative comparison with related work}. Note that previous works mostly focused on sequence level expressiveness and relied on subjective opinions from annotators and/or coders. In contrast, this work measures the expressiveness at video frame level using domain knowledge from affective computing. Also, to the best of our knowledge, there are no public visual affect datasets that were annotated at the video frame level for expressiveness. Considering this, a quantitative comparison with previous works is infeasible. Here, however, we discuss a qualitative comparison of the work from Uddin and Canavan \cite{uddin2021quantified}, which also computed an expressiveness score at video frame level. While this work is similar to ours, there are some differences including no bounds, the results are skewed towards the total number of AUs, and it lacks the ability to compare among different categories of expressions (e.g. happy, pain). The proposed algorithm addresses these limitations by providing an unbiased (toward $\#$ of AUs) lower and upper bounded expressiveness score. This is essential for a measurement scale, so we can perceive the relative importance of the expressiveness of a given frame.

\vspace{-2mm}
\subsection{Affective Behavior Analytics}
\label{sec:aff_behav_a}
\vspace{-3mm}
\subsubsection{\textbf{Granger causality between context and ground truth, and between context and QFE score}}
\label{sec:GrangerTestRes}
\vspace {-2mm}
We hypothesize that context will elicit facial expression since during data collection, in the BioVid pain dataset, temperature was used to elicit pain experience. We formulate this as a Granger causality (GC) test in which we use the temperature to test whether temperature Granger-causes facial expressiveness ($\tau$). We also tested the hypothesis that temperature Granger-causes the ground truth pain level. Note that temperature, pain level (\textbf{PL}), and QFE score $\tau$ for pain expression are modeled as temporal variables (Fig. \ref{fig:GC_testngEx}).

\textbf{Data preparation}. To test the hypotheses, we extracted one-minute-long video segments from the BioVid pain dataset, resulting in $1740$ video segments from $87$ subjects. Then, we computed the QFE score for pain expression ($\tau_{pain}$) using the AUs that are associated with the pain expression (AUs: $4, 6, 9, 10, 25$) \cite{werner2019automatic}. For each temporal variable, to make the variable stationary, we computed the difference between the consecutive values, and then, we performed an Augmented Dickey-Fuller (AD-Fuller) test \cite{ADF} to check whether the temporal variables are stationary or not, and found that all three variables passed the test. Then, we performed: $GC(temperature) \longrightarrow PL$ (i.e. temperature Granger-causes ground truth pain level); $GC(temperature) \longrightarrow \tau_{pain}$ (i.e. temperature Granger-causes facial pain expressiveness). We also performed an ablation study on the temporal history of the temperature using a lag ranged in between $[1, 10]$ seconds with an interval of $1$ second. For fair evaluation, we performed four different statistical tests: likelihood-ratio (\textbf{LR}) $\chi^{2}$ test, residual sum of squares (\textbf{SSR}) based $\chi^{2}$ test, parameters (\textbf{params}) F test, and SSR based F test. We reported the percentage of video segments that passed each test, separately and the percentage of segments that passed all four tests (\textbf{ALL}). It is important to note that we set the significance level $\alpha$ to $0.05$.


Table \ref{tab:GC_res} highlights the percentages of the video segments that passed the tests. We can infer from Table \ref{tab:GC_res} that in case of ALL, temperature Granger-caused the ground truth pain level (PL) in between $[47\%$, $100\%]$ of the video segments, while temperature Granger-caused the facial expressiveness in between $[10\%, 23\%]$ of the video segments. A lag of temperature in between $[5, 10]$ seconds was useful to predict the pain level and facial pain experience, which indicates applying heat for a longer time may induce a painful expression. As the temperature was self-calibrated by subjects for their own pain tolerance level, the strong predictive power of temperature towards PL is reasonable.

Per our hypothesis, we should observe high facial pain expressiveness when the temperature is high, however, based on the summary of $\tau_{pain}$ in Table \ref{tab:exprv_scr_summ} and the highlighted results in Table \ref{tab:GC_res}, we did not observe that in the collected affect. In Fig. \ref{fig:GC_testngEx}, we can see there is a strong relationship between pain level and temperature. However, even though we expect to observe facial pain expression with a change of temperature, we rarely observed that in this sequence. Considering this, analyzing expression on this dataset may not be reliable as it will give less insight into the pain level. Our results suggest that analyzing temperature could be a better solution towards perceiving pain (i.e. context is needed). Alternatively, this could be explained by inappropriate affect \cite{harris1956inappropriate}, where the subject's expression does not match the scenario. In our experiments, this would mean the subjects felt pain due to the temperature, however, they did not show a painful facial expression which can be validated by the construction theory of emotion \cite{barrett2017emotions}.

\vspace{-5mm}
\subsubsection{\textbf{Subjective Difference Analysis}}
\label{sec:sub_quant}
\begin{figure}
\centering
  \centering
  \includegraphics[width=0.99\linewidth]{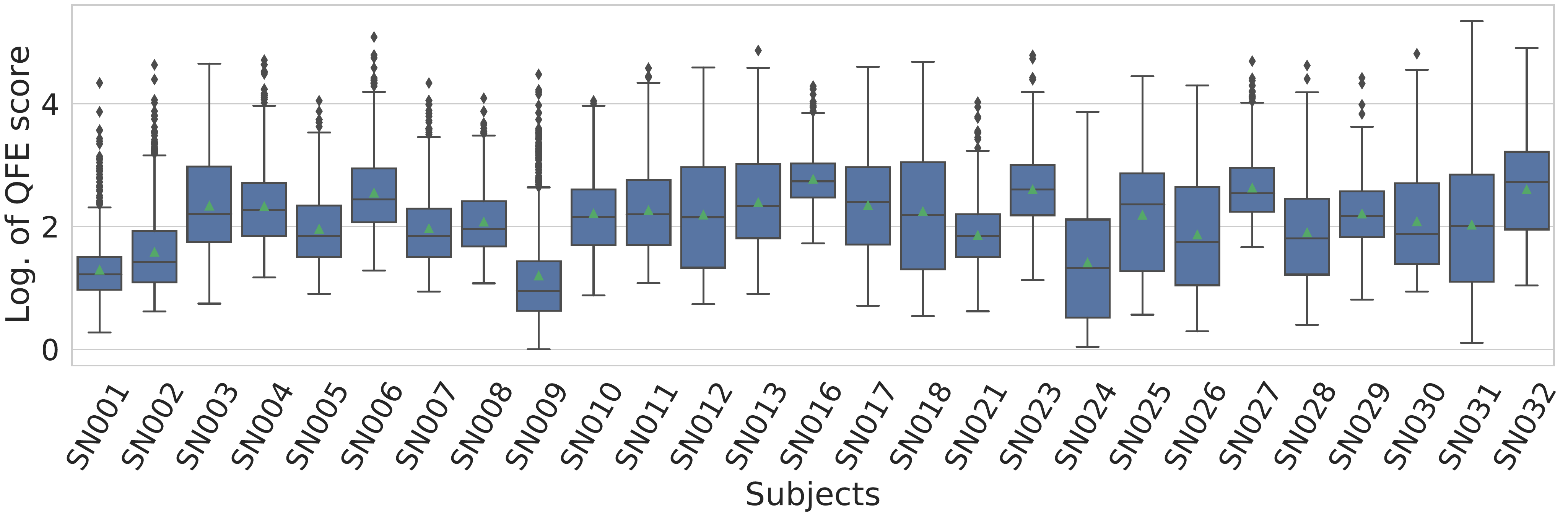}
\vspace{-3mm}
\caption{Facial expressiveness distribution across subjects.}
\label{fig:box_res_dist_summ_DISFA}
\vspace{-3mm}
\end{figure}

\vspace{-3mm}
We also conducted experiments to quantify subjectivity in terms of expression in context using DISFA. An example subjective difference is shown in Fig. \ref{fig:sampSeqSubDiff} from which we can observe that SN002 is more expressive than SN001, and SN003 is more expressive than SN002. The subject-specific distribution of the natural logarithm of the computed overall expressiveness score $\tau$ on DISFA is shown in Fig. \ref{fig:box_res_dist_summ_DISFA}. Based on the individual expressiveness distribution, we can say that people are quite different in terms of expressing themselves even though the context was the same.

\begin{figure}
\centering
\begin{subfigure}{.11\textwidth}
  \centering
  \includegraphics[width=.99\linewidth]{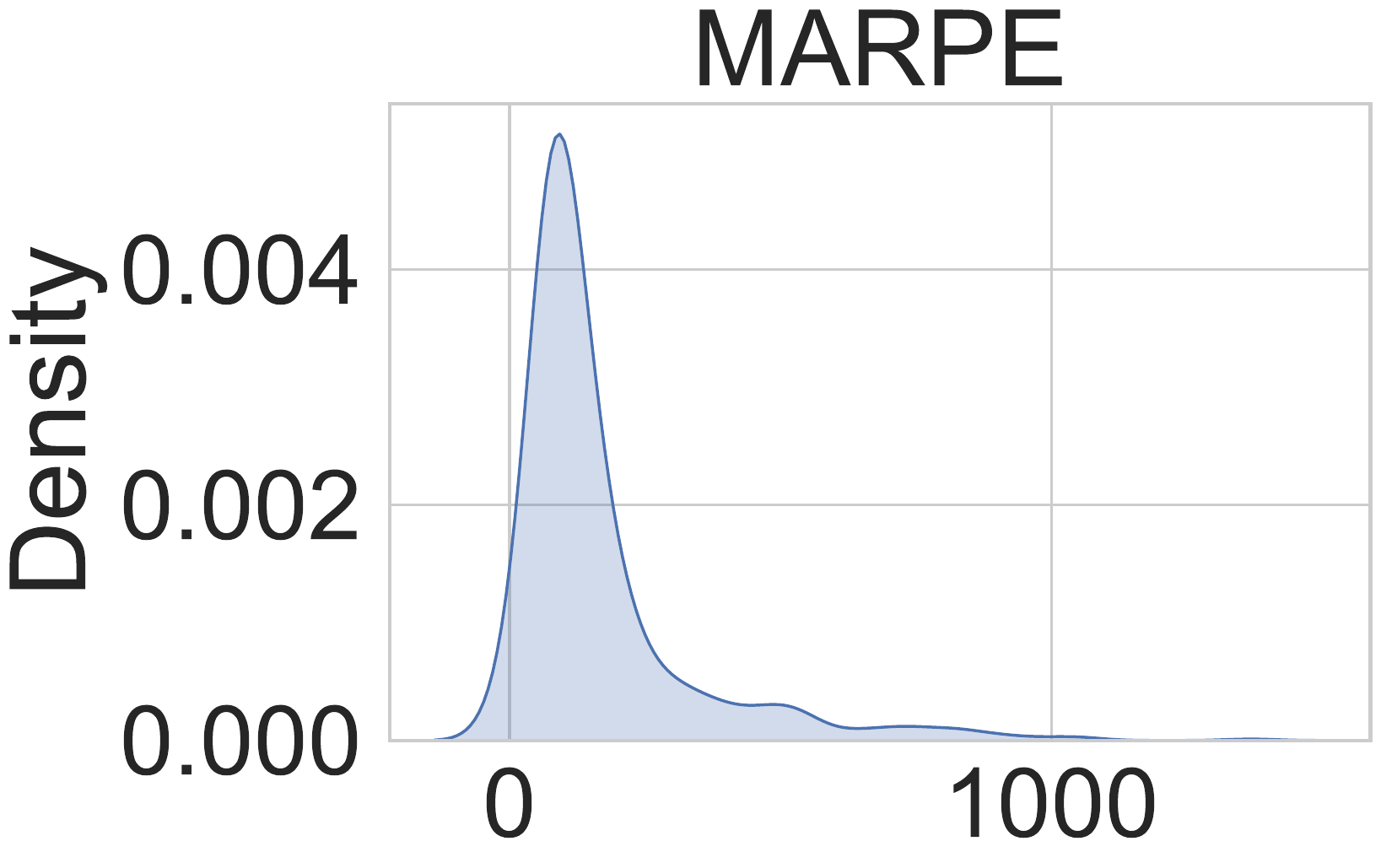}
  \caption{MARPE}
  \label{fig:MARPE_res_KDE_diff_dist_DISFA}
\end{subfigure}%
\begin{subfigure}{.11\textwidth}
  \centering
  \includegraphics[width=.99\linewidth]{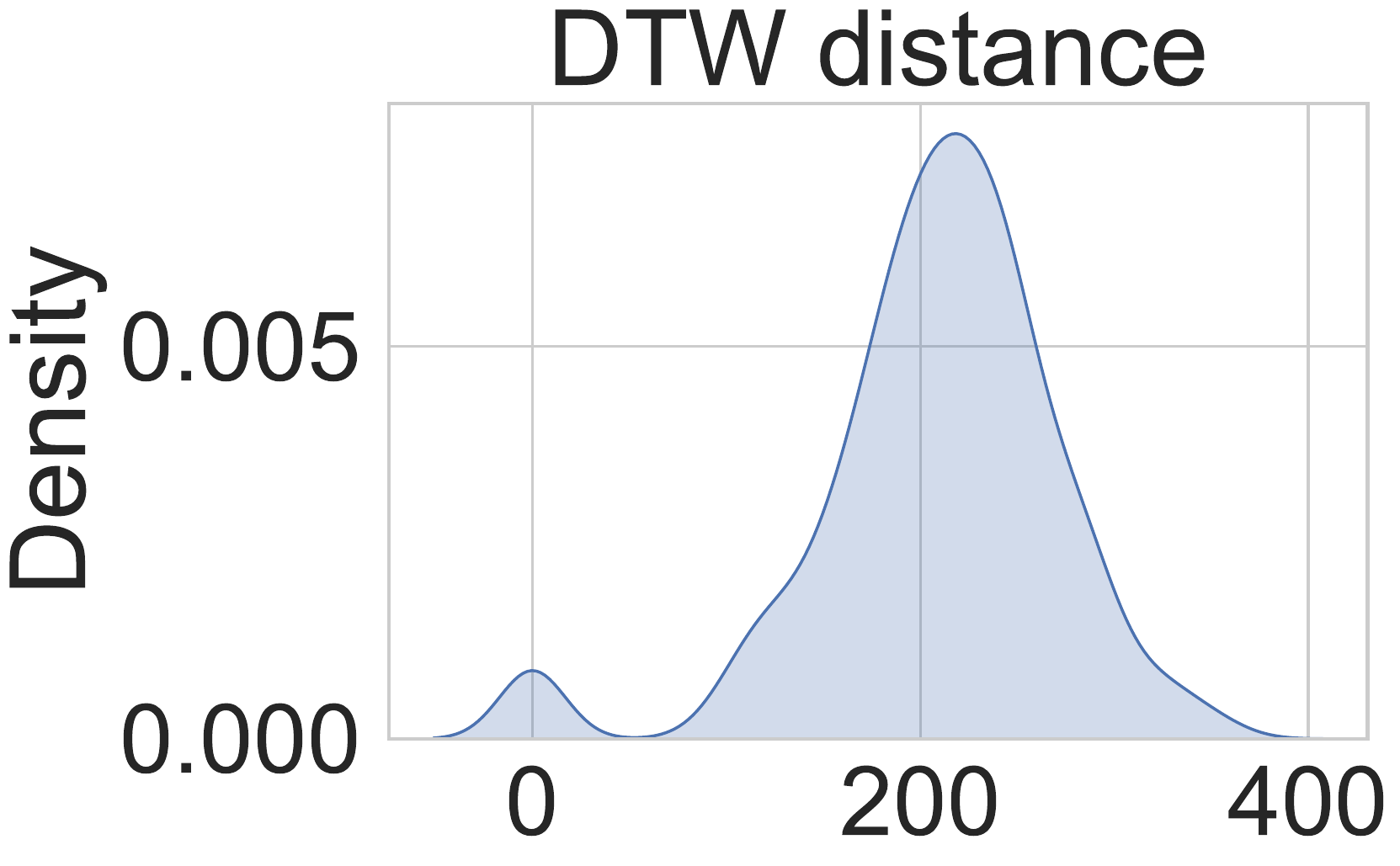}
  \caption{DTW dist.}
  \label{fig:dtw_res_KDE_diff_dist_DISFA}
\end{subfigure} 
\begin{subfigure}{.11\textwidth}
  \centering
  \includegraphics[width=.99\linewidth]{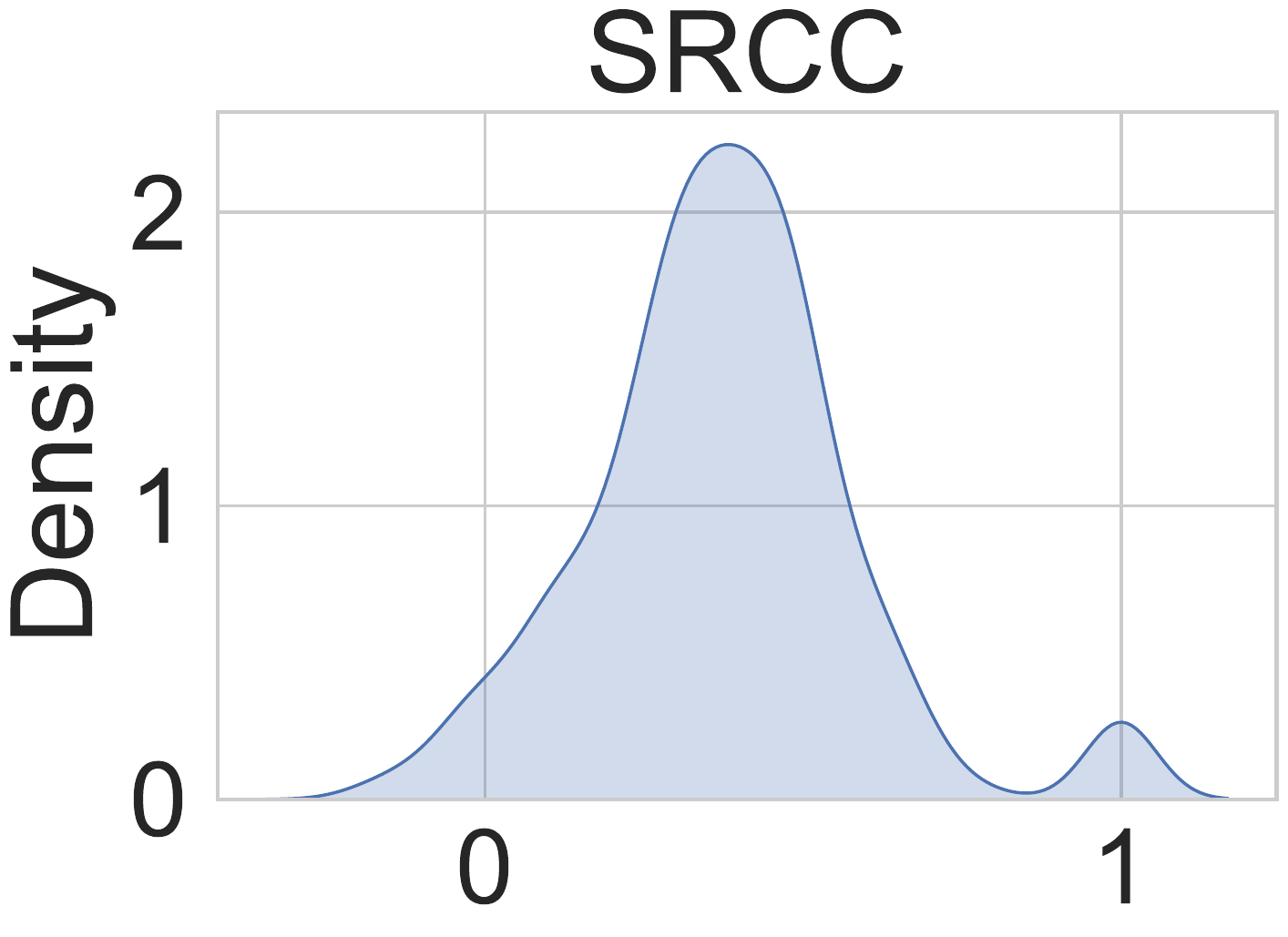}
  \caption{SRCC}
  \label{fig:srcc_res_KDE_diff_dist_DISFA}
\end{subfigure} 
\begin{subfigure}{.11\textwidth}
  \centering
  \includegraphics[width=.99\linewidth]{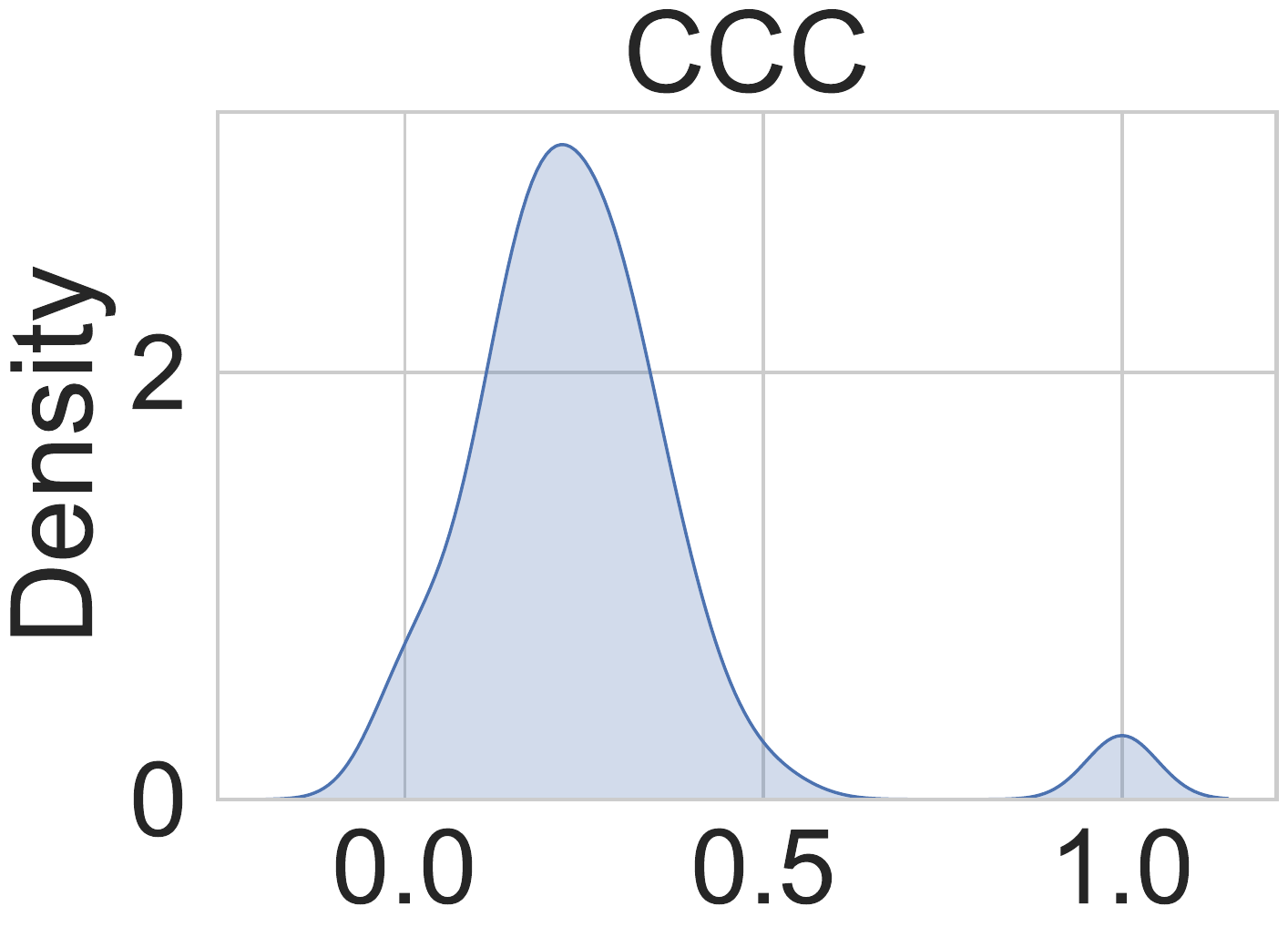}
  \caption{CCC}
  \label{fig:ccc_res_KDE_diff_dist_DISFA}
\end{subfigure} 
\vspace{-4mm}
\caption{Distribution of subjective differences. 
}
\label{fig:KDE_subDiff}
\vspace{-2mm}
\end{figure}

We measured subjective differences using four metrics: mean absolute-relative percentage error (\textbf{MARPE}), dynamic time warping (\textbf{DTW}) distance \cite{muller2007dynamic}, SRCC, and concordence correlation coefficient (\textbf{CCC}) \cite{liao2000note}. MARPE is defined as $MARPE = \frac{1}{n_f} \sum_{i=1}^{n_f} \abs{\frac{x - y}{x}} * 100$, where $x$ and $y$ are QFE scores computed from two subjects, and $n_f$ is the number of frames. We computed MARPE, DTW distance, SRCC, and CCC across all subjects for all combinations. Then, using the KDE \cite{silverman1986density} method, we estimated the distribution of the quantified subjectivity for each metric (Fig. \ref{fig:KDE_subDiff}). We also computed the mean and standard deviation (SD) of each metric, and obtained MARPE $ = 122.4\% \pm 130\%$, DTW distance $ = 93.13\pm 35.24$, SRCC $= 0.43\pm0.22$, and CCC $= 0.35\pm0.2$.  Here, high SD indicates high subject variability in terms of expressiveness.

\begin{figure}
\centering
\begin{subfigure}{.11\textwidth}
  \centering
  \includegraphics[width=.99\linewidth]{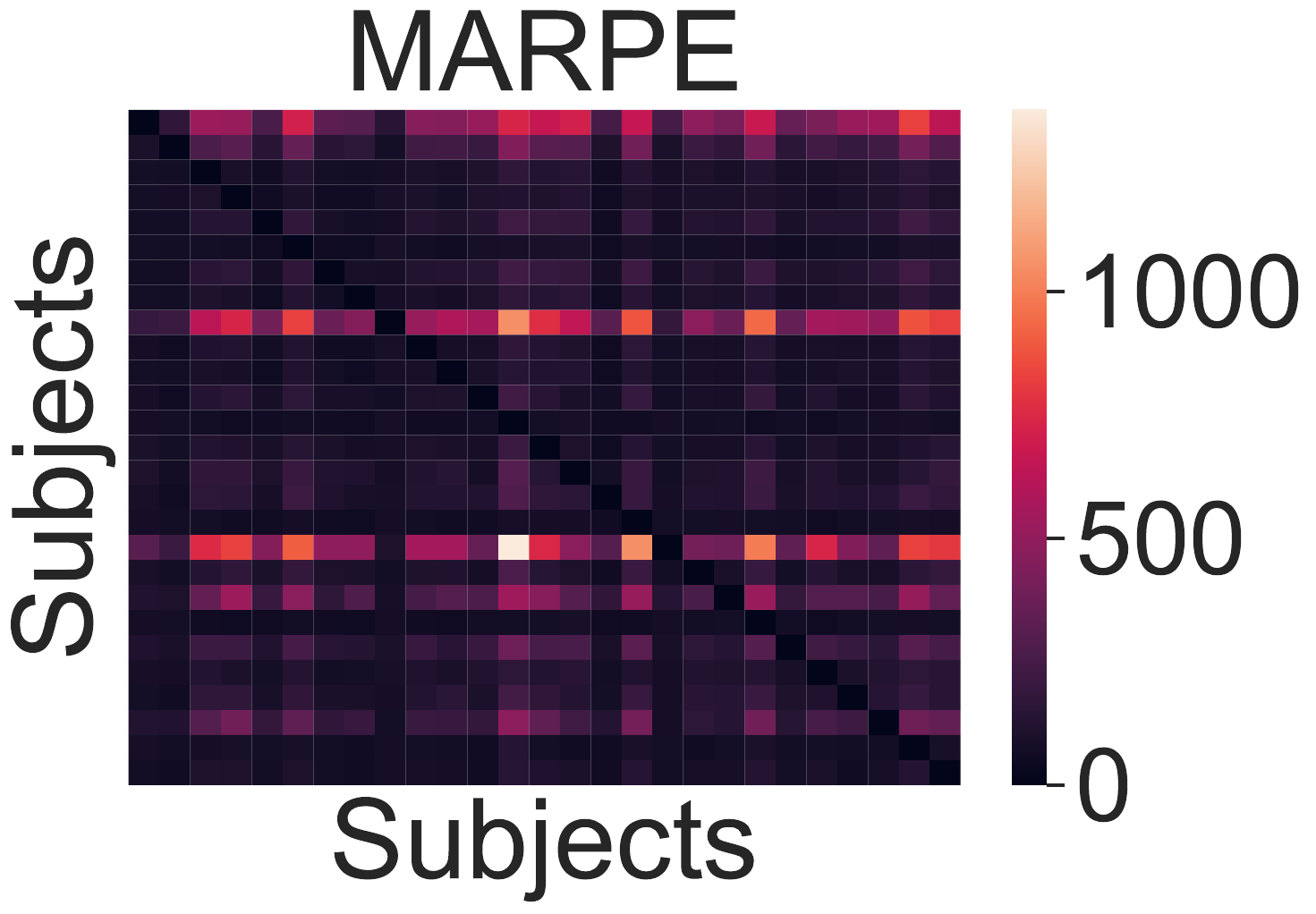}
  \label{fig:MARPE_res_heatMap_DISFA}
\end{subfigure}%
\begin{subfigure}{.11\textwidth}
  \centering
  \includegraphics[width=.99\linewidth]{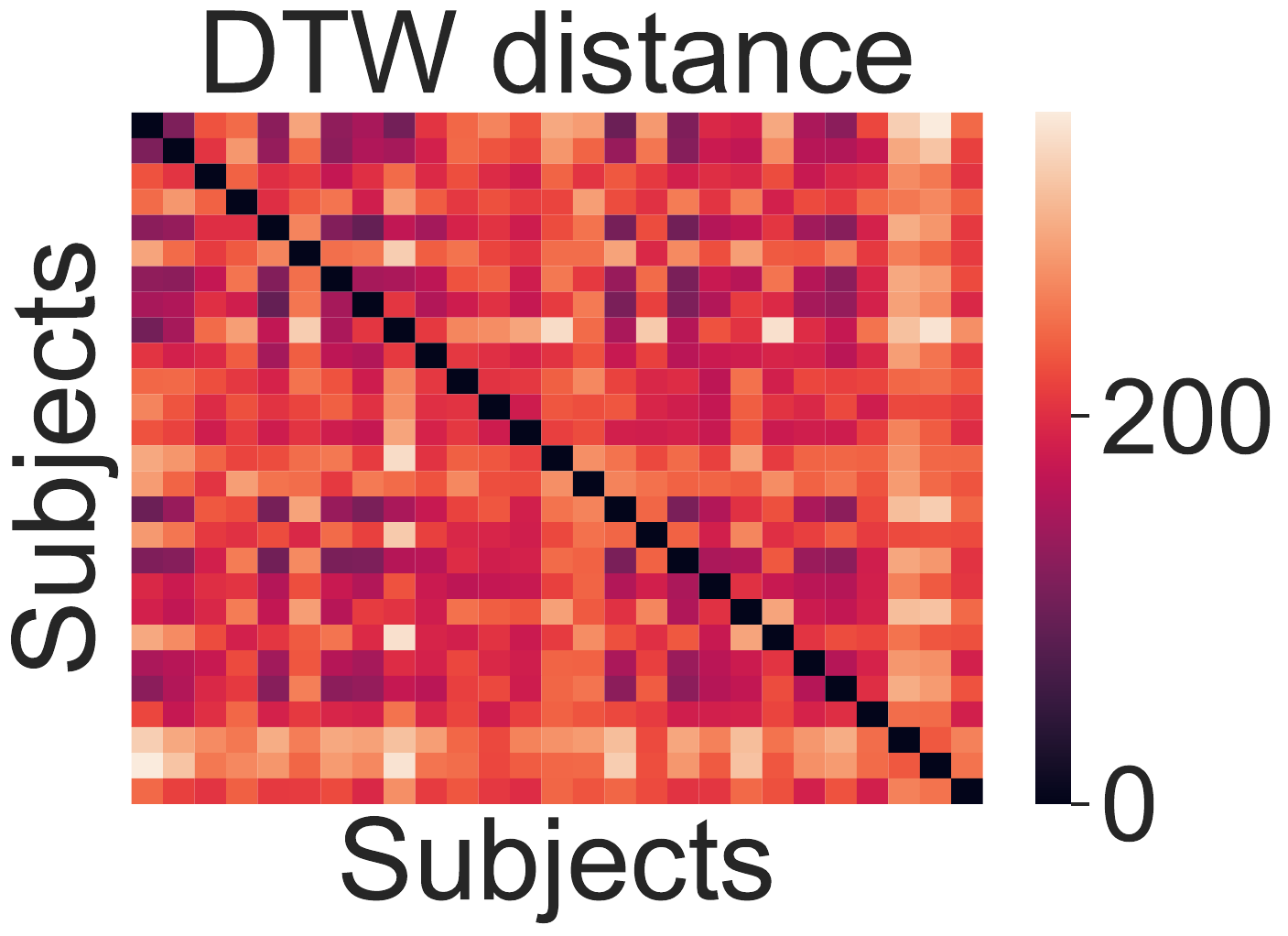}
  \label{fig:dtw_res_heatMap_DISFA}
\end{subfigure} 
\begin{subfigure}{.11\textwidth}
  \centering
  \includegraphics[width=.99\linewidth]{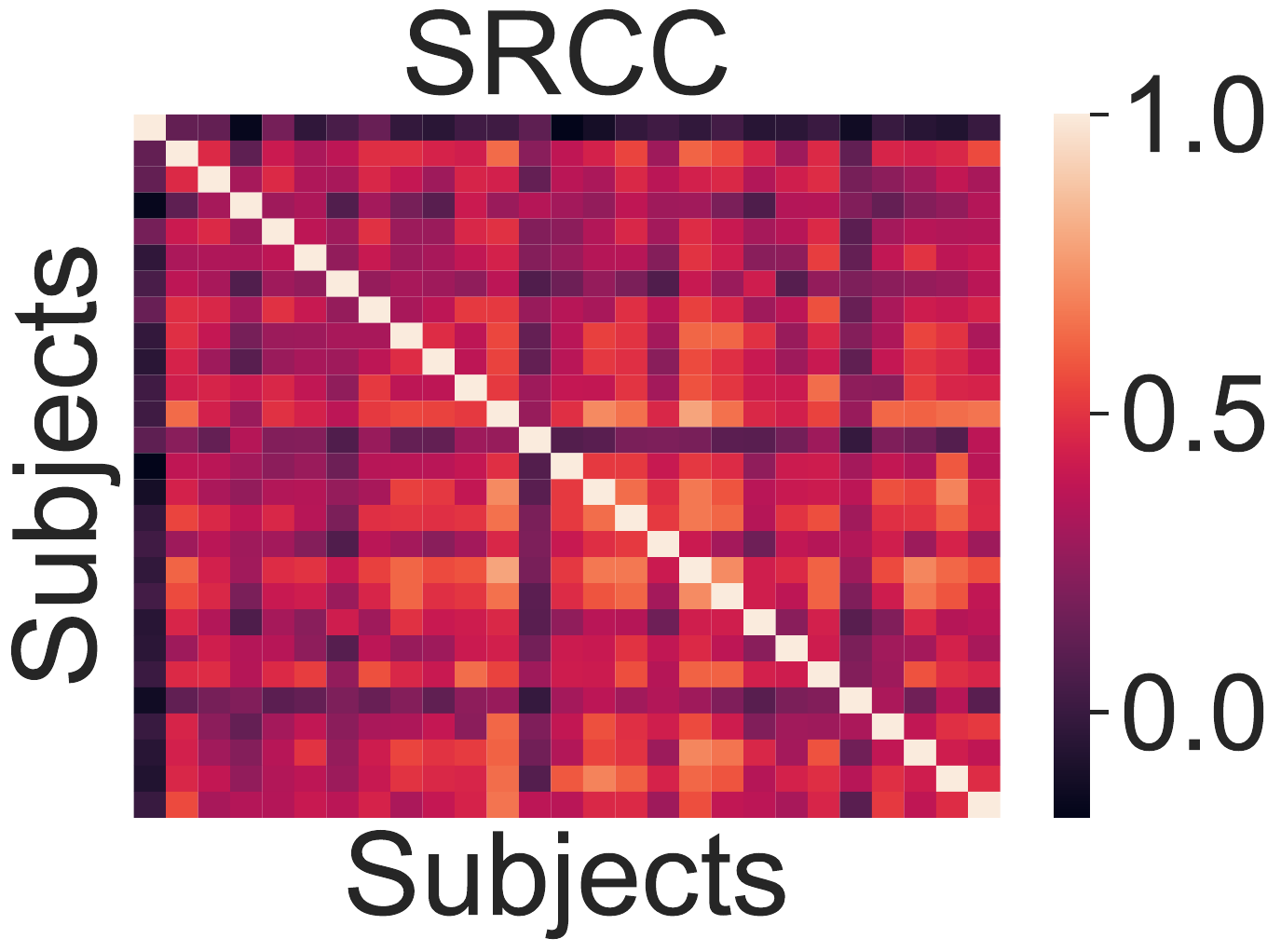}
  \label{fig:srcc_res_heatMap_DISFA}
\end{subfigure} 
\begin{subfigure}{.11\textwidth}
  \centering
  \includegraphics[width=.99\linewidth]{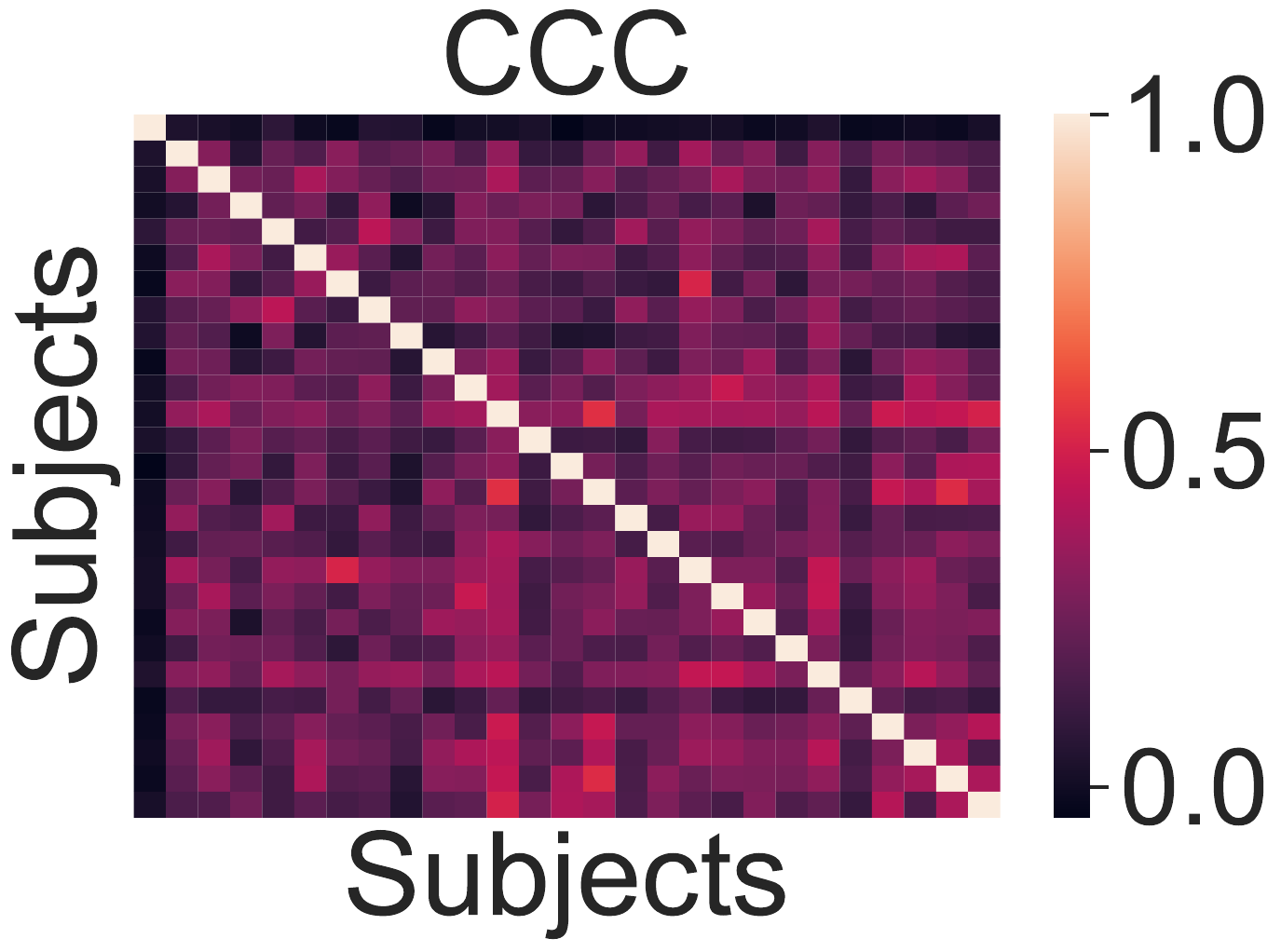}
  \label{fig:ccc_res_heatMap_DISFA}
\end{subfigure} 
\vspace{-7mm}
\caption{Heatmap representation of the quantified subjective differences by cross-referencing subjects in DISFA dataset. Here, \textit{higher MARPE, higher DTW distance, lower SRCC, lower CCC indicate higher subjective difference}. 
}
\label{fig:heatmapSubDiff}
\vspace{-5mm}
\end{figure}

In Fig. \ref{fig:heatmapSubDiff}, a cross-reference among subjects in terms of MARPE, DTW distance, SRCC, and CCC is shown. From the MARPE, we can infer that some subjects were more expressive than others, with a high margin. From SRCC and CCC, we can say that some subjects had high similarity of expressiveness compared to the others. Notice that CCC score is comparatively lower which can be explained, in part, as CCC looks for consistency in addition to the similarity in temporal sequences. It is important to evaluate these different metrics, as they conveyed different information (see Fig. \ref{fig:heatmapSubDiff}).
Based on our observation, subjects were not only different in terms of expressiveness but there were also differences in terms of lag and delay. To be precise, some subjects begin their expressions earlier, and had a longer duration, while others did not. For example, subject SN001, in DISFA (Fig. \ref{fig:sampSeqSubDiff}), was noticeably different from the rest of the subjects. When frame IDs were in between $[1000, 1100]$, subject SN002 was likely to be surprised and shocked while subject SN003 was likely to be confused, and subject SN001 was likely to be neutral in terms of expressions. Hence, we measured statistical significance along with the SRCC, and we found that SN001 was correlated with negative to random chance to subjects SN006, SN023, SN021, SN007, SN009, SN010, SN011, SN024, SN012, SN025, SN026, SN027, SN028, SN030, and SN031. In addition, SN029 was significantly different from the SN005 and SN016. The rest of the subjects ($25$) showed moderately positive to strong similarity ($p-value < 0.05$), indicating that context (stimuli) was effective at inducing affective facial expressions (Figs. \ref{fig:DISFA_subResponded_to_stimuli} and \ref{fig:srcc_res_KDE_diff_dist_DISFA}). To the best of our knowledge, this is the first work to show these findings. We encourage the use of them as a baseline for the quantification of subjectivity of facial expressions in DISFA.

\begin{figure}
\centering
\begin{subfigure}{.3\textwidth}
  \centering
  \includegraphics[width=.99\linewidth]{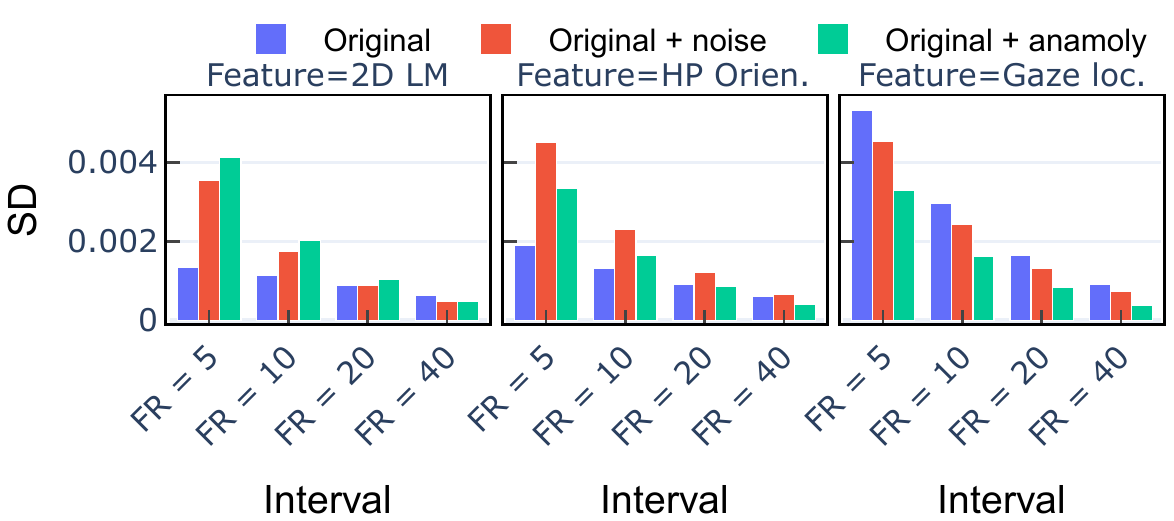}
  \vspace{-3mm}
  \caption{$\delta$ }
  \label{fig:qfe_scr_FA}
  \vspace{-3mm}
\end{subfigure} 
\begin{subfigure}{.09\textwidth}
  \centering
  \includegraphics[width=.99\linewidth]{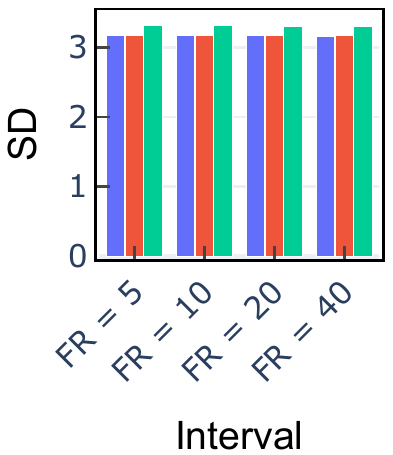}
  \vspace{-3mm}
  \caption{$\sigma$}
  \label{fig:qfe_phy_df}
  \vspace{-3mm}
\end{subfigure}%
\caption{Simulating influence of noise and anomaly incorporated from automated feature extraction models. ($SD=$ standard deviation).
} 
\label{fig:noise_simulation}
\vspace{-6mm}
\end{figure}

\vspace{-5mm}
\section{Discussion, Limitations and Future Work}
\label{sec:disc}
\vspace{-3mm}


An interesting direction, of this work, is the incorporation of dimensional models of expressions (e.g., valence, arousal) \cite{barrett1998discrete, blank2020emotional}. This could give us a better view of the expressiveness. Along with this, while OpenFace was used for feature extraction, the proposed approach is not limited to this, as other methods can be used such as AFAR \cite{ertugrul2019afar}. An investigation into which automated tool is best could be beneficial to the field, as automated prediction of features could introduce noise and anomalies into QFE. To evaluate this, we also simulated the influence of the presence of noise and anomalies on $\delta$ and $\sigma$. More formally, $D_{ns} = D*(1 + ns); ns \longleftarrow random(0, 0.05)$; where $D$ is the original data, and $ns$ is the noise generated from normal distribution, and $D_{ns}$ is generated noisy data (original + noise). To simulate anomalies, $D_{0.02a} = D_{0.02} * a; a \longleftarrow random(0, 2)$; we made only $2\%$ of the sample anomalous, and replaced those $2\%$ original samples with $D_{0.02a}$ to get anomalous data (original + anomaly) $D_a$. As can be seen in Fig. \ref{fig:noise_simulation}, noise and anomalies negatively influence the QFE scores. A possible way to mitigate this limitation is pre-processing the extracted features before computing QFE. For instance, in case of $\sigma$, the influence of outliers can be reduced using the fact that $0 <= AU_{intensities} <= 5$. To handle noise and outliers, we can leverage the confidence of the feature extraction models to deal with poorly extracted features. For instance, OpenFace outputs face detection probabilities, which can be used to pre-process data and mitigate noise and anomalies. Along with this, anomaly detection techniques can be used \cite{chandola2009anomaly}.



\vspace{-5mm}
\section*{Acknowledgment}
\vspace{-3mm}
We thank Nasimul Hasan and Liza Jivnani for providing human ratings. We also thank reviewers for their valuable feedback.
{\small
\bibliographystyle{ieee_fullname}
\bibliography{refs}
}
\clearpage

\end{document}